\documentclass[preprint,12pt]{elsarticle}

\usepackage{amssymb}
\usepackage{amsmath}
\usepackage{amsfonts}
\usepackage{algorithm}
\usepackage{algorithmic}
\usepackage{float}
\usepackage{graphicx}
\usepackage{booktabs} 
\usepackage{subcaption}
\usepackage{xcolor}
\usepackage{multirow}

\journal{arXiv}

\begin{document}

\begin{frontmatter}

\title{Time-Vertex Machine Learning for Optimal Sensor Placement in Temporal Graph Signals: Applications in Structural Health Monitoring}

\author[inst1]{Keivan Faghih Niresi}
\author[inst1]{Jun Qing}
\author[inst1]{Mengjie Zhao}
\author[inst1]{Olga Fink}

\affiliation[inst1]{{Intelligent Maintenance and Operations Systems Lab., EPFL, Lausanne, Switzerland}}

\begin{abstract}

Structural Health Monitoring (SHM) plays a crucial role in maintaining the safety and resilience of infrastructure. As sensor networks grow in scale and complexity, identifying the most informative sensors becomes essential to reduce deployment costs without compromising monitoring quality. While Graph Signal Processing (GSP) has shown promise by leveraging spatial correlations among sensor nodes, conventional approaches often overlook the temporal dynamics of structural behavior. To overcome this limitation, we propose Time-Vertex Machine Learning (TVML), a novel framework that integrates GSP, time-domain analysis, and machine learning to enable interpretable and efficient sensor placement by identifying representative nodes that minimize redundancy while preserving critical information. We evaluate the proposed approach on two bridge datasets for damage detection and time-varying graph signal reconstruction tasks. The results demonstrate the effectiveness of our approach in enhancing SHM systems by providing a robust, adaptive, and efficient solution for sensor placement.

\end{abstract}


\end{frontmatter}

\section{Introduction}
\label{sec:introduction}

Structural Health Monitoring (SHM) is a crucial technology for maintaining the safety, reliability, and functionality of civil, mechanical, and aerospace infrastructure. Through the continuous monitoring and analysis of structural responses to operational and environmental stresses, SHM systems can detect early signs of damage, enabling timely interventions that prevent catastrophic failures and extend the lifespan of structures \cite{coracca2023unsupervised}. The growing complexity of modern infrastructure and the increasing demand for resilience have amplified the importance of robust SHM systems. Civil infrastructures are constantly exposed to environmental and mechanical stressors, making them vulnerable  to damage from temperature variations, humidity, wind, seismic activity, and sustained or dynamic loads \cite{sun2020review,svendsen2022data, filizadeh2024risk}. Over time, such factors introduce deterioration of structural elements that may lead to the risk of collapse if the process is not detected in due time \cite{zhang2022causes}.

Recent advances in sensor technology and the Internet of Things (IoT) have enabled extensive SHM networks for bridges, buildings, dams, tunnels, and other critical structures \cite{peng2023development, abdelraheem2022design, he2022integrated, mishra2022structural}. These networks use strain gauges, accelerometers, temperature, and displacement sensors to continuously monitor structural responses such as stress, vibration, and torsion, together with environmental influencing factors. Data are transmitted to central systems where domain experts analyze responses, detect anomalies, and predict potential failures, enabling proactive maintenance \cite{fink2020potential, liu2024structural}.

Effective SHM depends critically on the strategic placement of sensors, as sensor locations directly determine the quality, reliability, and coverage of collected data \cite{hassani2023systematic}. In practice, optimal sensor placement ensures that critical structural regions such as high-stress zones or vibration-sensitive areas are continuously monitored while minimizing installation and maintenance costs \cite{malings2018value}. Efficient sensor placement, often termed sampling set selection, allows engineers to capture essential structural responses with a reduced number of sensors, thereby lowering system complexity and data management without compromising monitoring accuracy \cite{suslu2023understanding}. This is particularly important for large or complex structures, where dense sensor networks are impractical due to cost, power limitations, and accessibility constraints \cite{chen2024wireless}. For structural engineers, this translates to a fundamental question: what is the minimum set of sensors that provides unambiguous data for assessing global structural integrity and localizing potential damage?

Traditionally, finite element (FE) models and engineering intuition \cite{zhang2023data} have guided sensor placement by identifying structurally sensitive regions such as high-stress or high-modal-participation zones. While these methods remain essential for determining plausible sensor locations, they typically do not provide a systematic way to optimize the number or configuration of sensors. As a result, traditional model-based approaches may lead to redundant or non-optimal layouts, even though they still play a critical role in establishing the initial candidate set. To address this specific limitation, data-driven and optimization-based methods have been developed.

Information-theoretic approaches, including mutual-information, entropy-based, and Fisher-information formulations, seek to maximize the informativeness of selected sensor locations \cite{guestrin2005near, krause2008near, ly2017tutorial}. In particular, Fisher-information–based optimal design employs A-, D-, E-, and T-optimality criteria to quantify the precision gained from different sensor configurations \cite{rissanen1996fisher, pukelsheim2006optimal, chen2021cramer}. However, these methods commonly rely on simplified structural assumptions and may neglect spatial correlations, limiting their effectiveness for distributed physical systems \cite{kim2024effective, sahu2022optimal}.

Heuristic approaches such as greedy algorithms or sparse optimization provide practical alternatives for large-scale structures. These methods iteratively select sensors to balance computational efficiency with near-optimal coverage. However, because they rely on locally optimal decision rules and often do not explicitly account for spatial correlations, the resulting sensor configurations may deviate from the true global optimum in complex or distributed systems.

Recently, graph-based approaches have emerged as a powerful tool for practical SHM \cite{bloemheuvel2021computational, cheema2024computationally}. By representing sensors as nodes and structural interactions as edges, these methods capture spatial dependencies in a way that can mirror an engineer's understanding of load paths and structural connectivity, without requiring strict probabilistic assumptions. Graph signal processing (GSP) techniques further leverage this structure to guide sensor selection, effectively identifying critical nodes in the network \cite{shuman2013emerging, leus2023graph}. However, many GSP-based methods overlook temporal variations in structural behavior, limiting their applicability for time-varying responses.

To address this limitation, we propose Time-Vertex Machine Learning (TVML), a novel framework that integrates GSP, temporal signal processing, and traditional machine learning to design an interpretable sensor placement strategy. TVML provides a data-driven rationale for structural engineers by grouping locations with similar dynamic behavior and identifying the most informative sensor within each group. The key novelty of the proposed approach lies in modeling sensor signals jointly across time and graph domains, enabling a unified framework that captures time, spectral, and spatial domains. The TVML framework begins by extracting meaningful statistical and spectral features to characterize sensor behavior and reveal  patterns that facilitate efficient  clustering. These clusters group sensor  signals with similar behaviors, reducing redundancy while preserving the  fidelity of the system representational. Within each cluster, GSP and graph-theoretic principles are applied to identify critical sensors that balance information richness with redundancy minimization. This approach  ensures that the selected sensors are both interpretable and functionally significant within the network, leading to optimized sensor placement for SHM applications. After selecting sensors through  the proposed TVML method, we applied Spatial-Temporal Graph Neural Networks (STGNN) \cite{jin2024survey} to the resulting graph data, utilizing only  the selected subset of nodes. We  evaluated its performance on two downstream tasks: damage detection and time-varying graph signal reconstruction, using two datasets based on real-world measurements  and numerical simulations. The results  demonstrate  the adaptability and effectiveness  of our method in optimizing sensor placement,   enhancing system performance, and enabling cost-effective SHM. 

This paper can be summarized as follows:

\begin{itemize}
\item We propose TVML, a novel framework that combines GSP, temporal signal processing, and machine learning to develop  an interpretable sensor selection strategy.

\item We extract meaningful features to cluster sensors with similar behaviors. Within each cluster, GSP and graph-theoric principles are applied to identify  critical sensors that maximize  information richness while minimizing redundancy.

\item   We evaluate the proposed approach using STGNN  on two key tasks: damage detection and time-varying graph signal reconstruction for bridge infrastructure. The results demonstrate the effectiveness of TVML in optimizing sensor placement and enhancing SHM performance.
\end{itemize}

The remainder  of this paper is organized as follows: Section \ref{sec:preliminaries} provides a comprehensive review of the fundamentals of GSP, graph learning, and sensor selection. In Section \ref{sec:method}, we introduce the proposed TVML methodology for sensor selection and describe the STGNN architecture used for evaluation. Section \ref{sec:experiments} presents the case studies and results, highlighting the performance of our approach in comparison to several baseline methods. Finally, Section \ref{sec:conclusion} concludes the paper by summarizing our findings and discussing potential  directions for future research.

\section{Preliminaries}
\label{sec:preliminaries}
In this paper, sets are denoted  using calligraphic fonts (e.g., $\mathcal{S}$), with their cardinality represented  as $|\mathcal{S}|$. Vectors are expressed  using bold lowercase letters (e.g., $\mathbf{x}$), while matrices are represented by bold uppercase letters (e.g., $\mathbf{X}$). The transpose of a matrix $\mathbf{X}$ is denoted as  $\mathbf{X}^\mathsf{T}$, and the trace of a matrix $\mathbf{A}$ is written as $\mathsf{tr}(\mathbf{A})$. The notation $\mathsf{diag}(\mathbf{A})$ is used to extract the diagonal elements of a matrix $\mathbf{A}$ and  construct  a diagonal matrix from them. Additionally, $\|\cdot\|_p$ represents the $p$-norm, and $\odot$ denotes the Hadamard (element-wise) product.

\subsection{Graph Signal Processing} 

A graph $\mathcal{G} = (\mathcal{V}, \mathcal{E}, \mathbf{A})$ is defined by  a set of $N$ nodes (vertices) $\mathcal{V}$, an edge set $\mathcal{E}$, and a weighted adjacency matrix $\mathbf{A} \in \mathbb{R}^{N \times N}$. The degree of a node is calculated as the sum of the weights of all edges connected to it. The diagonal degree matrix $\mathbf{D}$ contains these node degrees along its diagonal, where each entry is defined as  $\mathbf{D}(i,i) = \sum_{j=1}^N \mathbf{A}(i,j)$. Using the degree matrix and the adjacency matrix, the graph Laplacian is defined as $\mathbf{L} := \mathbf{D} - \mathbf{A}$.

The graph Laplacian $\mathbf{L}$ is a positive semi-definite matrix and can therefore be decomposed using  eigenvalue decomposition as $\mathbf{L} = \mathbf{U} \mathbf{\Lambda} \mathbf{U}^\top$. In this decomposition, $\mathbf{U}$ is a matrix whose columns are the eigenvectors of $\mathbf{L}$, forming a Fourier basis for the graph, and $\mathbf{\Lambda}$ is a diagonal matrix containing  eigenvalues $\{\lambda_1, \dots, \lambda_N\}$, which are referred to as the graph frequencies.

A graph signal assigns  a real-valued measurement to each node in a graph and is represented as a vector $\mathbf{x} \in \mathbb{R}^N$, where $N$ is the number of nodes. A time-varying graph signal \cite{acar2022learning, qiu2017time}, also referred to as a temporal graph signal \cite{mcneil2021temporal}, captures  the evolution of these  signals over time across the graph's nodes. For example, data collected from a sensor network at multiple time instants can be modeled as a time-varying graph signal. A time-varying graph signal spanning $T$ time instants is expressed  as a matrix $\mathbf{X} = [\mathbf{x}_1, \mathbf{x}_2, \dots, \mathbf{x}_T] \in \mathbb{R}^{N \times T}$, where each column $\mathbf{x}_t \in \mathbb{R}^N$ represents the graph signal at time $t$.

The Graph Fourier Transform (GFT) of a graph signal $\mathbf{x} \in \mathbb{R}^N$ is defined as:
\begin{equation}
\mathcal{GFT}\{\mathbf{x}\} = \hat{\mathbf{x}} = \mathbf{U}^\mathsf{T} \mathbf{x},
\end{equation}
where $\mathbf{U}$ is the matrix of Fourier basis vectors, corresponding to the eigenvectors of the graph Laplacian $\mathbf{L}$. The GFT provides a spectral representation of the graph signal by projecting it onto the eigenvector space of the Laplacian. 
Filtering a graph signal can be expressed as:
\begin{equation}
\mathbf{y} = g(\mathbf{L})\mathbf{x} = \mathbf{U} g(\mathbf{\Lambda}) \mathbf{U}^\mathsf{T} \mathbf{x},
\end{equation}
where $\mathbf{x}$ is the input signal, $\mathbf{y}$ is the filtered  output signal, and $g(\lambda)$, is the filter function that defines the frequency response  of the filter. The operator $g(\mathbf{\Lambda})$ applies the function $g$ to the eigenvalues of $\mathbf{L}$ (i.e., the graph frequencies), enabling manipulation of the signal in the spectral domain.

\subsection{Learning Graph from Data}

Graph signal smoothness is a fundamental concept in GSP, often quantified using the Dirichlet form. The \(p\)-Dirichlet form measures the variation of a graph signal across connected nodes and is expressed as \cite{mateos2019connecting, montenegro2006mathematical, niresiigl2024}:
\begin{equation}
S_p(\mathbf{x}) = \frac{1}{p} \sum_{i \in \mathcal{V}} \|\nabla_i \mathbf{x}\|_2^p,
\end{equation}
where \(\|\nabla_i \mathbf{x}\|_2\) quantifies the signal variation across edges connected to node \(i\). For \(p=2\), this formulation reduces to the graph Laplacian quadratic form:
\begin{equation}
S_2(\mathbf{x}) = \sum_{(i, j) \in \mathcal{E}} \mathbf{A}(i, j) [\mathbf{x}(j) - \mathbf{x}(i)]^2  = \mathbf{x}^\mathsf{T}\mathbf{L}\mathbf{x},
\end{equation}
A smooth graph signal exhibits a low \(S_2(\mathbf{x})\), indicating  that connected nodes tend to have similar signal values. In the case of time-varying signals, represented as a series  of snapshots \(\mathbf{X} = [\mathbf{x}_1, \mathbf{x}_2, \dots, \mathbf{x}_T] \in \mathbb{R}^{N \times T}\), the concept of smoothness naturally extends  to the aggregated graph signal:
\begin{equation}
S_2(\mathbf{X}) = \sum_{i=1}^{T} \mathbf{x}_i^\mathsf{T} \mathbf{L} \mathbf{x}_i = \mathsf{tr}(\mathbf{X}^\mathsf{T} \mathbf{L} \mathbf{X}).
\end{equation}
Smoothness  is a fundamental principle for graph inference \cite{dong2016learning}. Building on this concept, \cite{pmlr-v51-kalofolias16} introduced a framework for inferring graphs directly from data by exploiting the smoothness property of graph signals. The objective function for this framework is defined as:
\begin{equation}
\begin{aligned}
\min_{\mathbf{A}} & \quad \|\mathbf{A} \odot \mathbf{Z}\|_{1,1} - \lambda_1 \mathbf{1}^\mathsf{T} \log(\mathbf{A1}) + \frac{\lambda_2}{2} \|\mathbf{A}\|_F^2 \\
\text{s.t.} & \quad \mathbf{A} \in \mathbb{R}_+^{N \times N}, \quad \mathbf{A} = \mathbf{A}^\mathsf{T}, \quad \mathsf{diag}(\mathbf{A}) = \mathbf{0}.
\end{aligned}
\label{eq:kalo}
\end{equation}
In this formulation, the term \(\|\mathbf{A} \odot \mathbf{Z}\|_{1,1}\) represents the elementwise \(\ell_1\)-norm, where  \(\mathbf{Z}\) represents the distance matrix based on signal differences, with entries \(\mathbf{Z}(i, j) = |\mathbf{x}(i) - \mathbf{x}(j)|^2\). This term emphasizes edges with larger signal differences, encouraging smoothness in the learned adjacency matrix $\mathbf{A}$. The logarithmic barrier term, \(-\lambda_1 \mathbf{1}^\mathsf{T} \log(\mathbf{A1})\), ensures positive node degrees while allowing edge weights to reach zero. The regularization term \(\frac{\lambda_2}{2} \|\mathbf{A}\|_F^2\) promotes numerical stability and sparsity. The constraints enforce that the learned adjacency matrix  \(\mathbf{A}\) is non-negative, symmetric, and without self-loops, respectively.

While the adjacency matrix in Eq.~(\ref{eq:kalo}) is learned purely from signal smoothness, the proposed TVML framework is intentionally designed to be modular and flexible. In practical SHM applications, where the physical topology or sensor layout is already known, the adjacency matrix \(\mathbf{A}\) can be directly defined from this spatial configuration and provided as an input to the framework, bypassing the graph-learning stage entirely. The graph-learning step is therefore not mandatory, but serves as a data-driven alternative for cases where the topology is not or only partially known, enabling the inference of functional connectivity between sensors. In addition, in many real-world deployments, some sensors are physically fixed at critical structural locations (e.g., supports or beams) and cannot be removed. These mandatory sensors can be pre-specified, while TVML can focus on optimizing the number and placement of additional flexible sensors.

\subsection{Graph Signal Sampling}
Graph signal sampling, closely tied to the concept of sensor placement, involves observing a subset of nodes in a graph and estimating the signal values at unobserved  nodes based on the observed signals. This process, referred to as vertex-domain sampling \cite{tanaka2018spectral}, is the graph analog of time-domain sampling. The selected nodes for observation form the sampling set, denoted as \( \mathcal{S}^* \subset \mathcal{V} \), where \( \mathcal{S}^* \) contains \( M < N \) nodes. After the sampling set \( \mathcal{S}^* \) is determined, the sampled signal can be utilized to approximate or reconstruct  bandlimited graph signals using advanced techniques \cite{narang2013signal}. However, in practical scenarios, because the cutoff frequency is typically unknown, the signal may not be strictly bandlimited, and observed samples are often affected by additive noise, making exact reconstruction infeasible. Mathematically, the sampled graph signal \( \mathbf{x}_s \in \mathbb{R}^M \) is expressed as:
\begin{equation}
\mathbf{x}_s = \mathbf{I}_{\mathcal{S}^*} \mathbf{x},
\end{equation}
where \( \mathbf{I}_{\mathcal{S}^*} \in \mathbb{R}^{M\times N} \) is a submatrix of the identity matrix \( \mathbf{I} \), containing only the rows corresponding to the indices of the sampling set \( \mathcal{S}^* \). The objective is to design \( \mathbf{I}_{\mathcal{S}^*}\) to maximize signal recovery performance.

\section{Methodology}
\label{sec:method}

\subsection{Problem Definition}

Let $\mathcal{S} = \{s_1, s_2, \dots, s_N\}$ denote  a set of $N$ sensors, each providing time-series data  that captures information about the structure’s state. The objective is to identify a subset $\mathcal{S}^* \subset \mathcal{S}$ with $|\mathcal{S}^*| = M < N$, such that the selected sensors in $\mathcal{S}^*$ are the most informative for the given task. The sensor network is modeled  as a graph $\mathcal{G} = (\mathcal{V}, \mathcal{E}, \mathbf{A})$, where each node $v_i \in \mathcal{V}$ corresponds to a sensor, and the edges $\mathcal{E}$ represent spatial or functional relationships among sensors.   The adjacency matrix \(\mathbf{A}\) encodes the strength of these relationships, providing a structured framework for analyzing the sensor network and guiding the selection of the optimal subset.

\subsection{Proposed Method}

\begin{figure*}
    \centering
    \includegraphics[width=\linewidth]{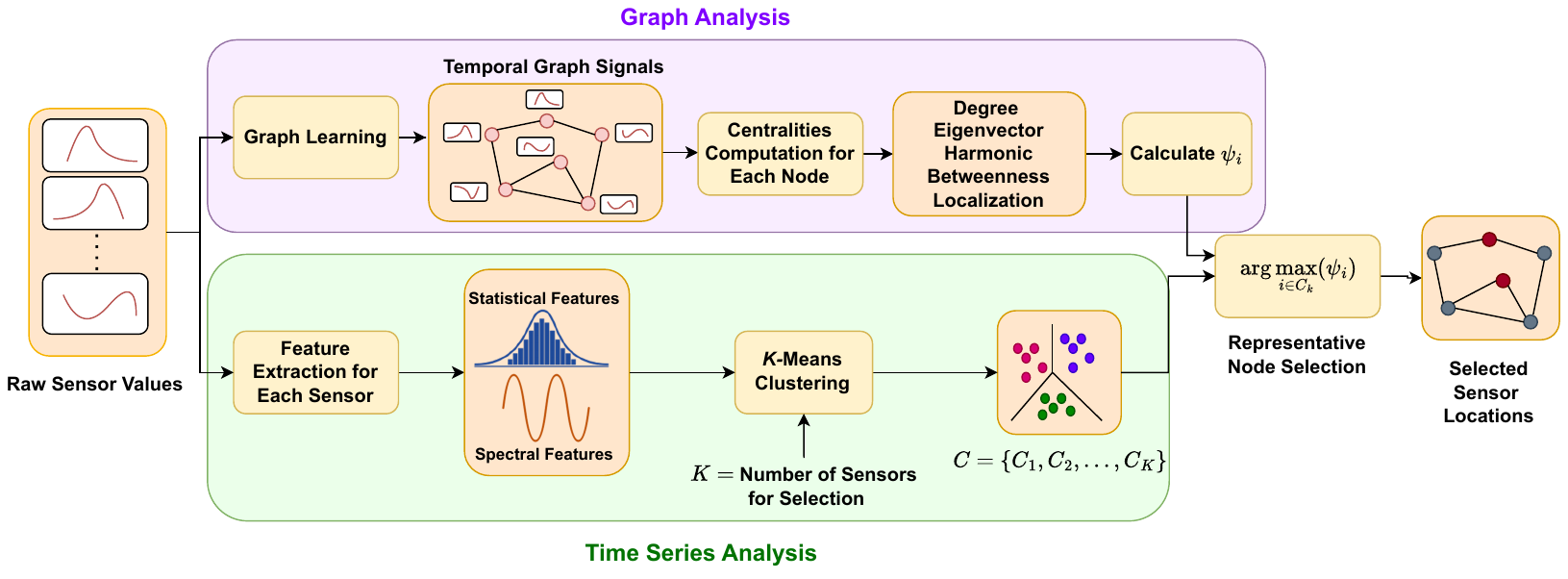}
    \caption{Block Diagram of the Proposed TVML Framework for Sensor Selection. The framework combines time series and graph-based analysis to select informative and non-redundant sensors. Statistical and spectral features guide clustering, while graph centrality measures help identify representative sensors within each cluster.}
    \label{fig:TVMLDiagram}
\end{figure*}

We propose a novel sensor selection methodology to identify the most informative and least redundant sensors for SHM. Unlike traditional approaches that rely purely on spatial configurations or raw signal measurements, our method leverages interpretable statistical and (graph) signal processing tools to guide sensor grouping and selection. The proposed methodology for selecting the most informative sensors from  a set of $N$ available sensors with defined locations in the graph involves three main steps as illustrated in Figure \ref{fig:TVMLDiagram}: (1) feature extraction, (2) clustering, and (3) representative sensor selection within clusters. These steps systematically reduce the number of sensors to a smaller  subset $M$ ($M < N$), while preserving  as much of the original information as possible. (1) The feature extraction step aims to capture statistical and spectral characteristics of each sensor's measurements. For each sensor $s_i$, its time-series data  is represented as $x_i(t)$, where $t$ denotes  the time index. A set of statistical features is computed from each sensor’s time series, quantifying the unique patterns and behaviors of the sensor readings such as variance, mean, skewness, and kurtosis.  These features provide  a compact yet informative representation of the sensor's measurements, facilitating comparison during the clustering process.  By working with these extracted features instead of  raw time-series data, we reduce the complexity of subsequent analyses while retaining key information about each sensor’s behavior. (2) Following  feature extraction, the sensors are grouped into $M$ clusters, where each cluster contains sensors with similar statistical characteristics. The clustering is performed  using the $K$-means algorithm, with the number of clusters set to $K = M$, corresponding to the desired number of selected sensors. The $K$-means algorithm partitions the sensors by minimizing the variance within each cluster, creating  groups of sensors with similar data profiles. This step reduces redundancy among sensor, as as sensors in the same cluster tend to exhibit highly correlated behavior. (3) Once the clusters are formed, a representative sensor is selected from each cluster to capture the essential characteristics of the group. To identify the most informative representative,  each sensor’s significance within its cluster  is evaluated based on its centrality in a sensor graph. This graph can be constructed using  spatial or physical sensor locations, or based on the similarity of signal behaviors if no spatial layout information is available. Sensors with higher centrality in the graph are considered more representative of their cluster, ensuring that the selected subset retains the key information from the original sensor network.

\subsubsection{Feature Extraction for Sensor Signals}
Each sensor $s_i$ generates a time-series signal $x_{i}(t)$ at time $t$. To capture the essential characteristics of each sensor’s signal, we define a feature vector $\mathbf{f}_i \in \mathbb{R}^d$ , comprising both statistical and spectral properties. These features are designed to effectively  distinguish sensor behavior, enabling the selection of diverse and informative sensors.

\emph{Statistical Features}: We compute statistical features to characterize the signal, including the mean (\(\mu\)), maximum (\(\max\)), minimum (\(\min\)), variance (\(\sigma^2\)), skewness (\(\gamma_1\)), kurtosis (\(\gamma_2\)), and root mean square (RMS). These features capture different aspects of signal behavior, such as central tendency, spread, asymmetry, and intensity. Detailed mathematical formulations are provided in the Appendix.

\emph{Spectral Features}: Spectral features provide insights into the frequency characteristics of the signal. Using the Fast Fourier Transform (FFT), we  transform the time-domain signal \( x(t) \) into the frequency domain:

\begin{equation}
X(f) = \sum_{t=0}^{T-1} x(t) \cdot e^{-j 2\pi \frac{f t}{T}}, \quad f = 0, 1, \dots, T-1
\end{equation}
where \( X(f) \) is the complex-valued FFT at frequency bin \( f \). The magnitude of the FFT, which represents the signal’s energy at different frequencies, is computed as:

\begin{equation}
|X(f)| = \sqrt{\Re(X(f))^2 + \Im(X(f))^2}
\end{equation}

From the FFT, we extract the following features:

\emph{Magnitude of the First $k$ Frequency Bins (MB)}: The magnitude at the first 
$k$ frequency bins captures the energy distribution in low-frequency bands:
\begin{equation}
\mathsf{MB} = \left[ |X(f_1)|, |X(f_2)|, \dots, |X(f_k)| \right]
\end{equation}
where \( k \) is the number of frequency bins we are considering. These values represent the strength of the signal at different frequency bands. 

\emph{Peak Frequency $f_{\text{peak}}$}: This represents the frequency with the maximum magnitude:
\begin{equation}
f_{\text{peak}} = f_{k_{\text{peak}}}, \quad \text{where} \quad k_{\text{peak}} = \arg\max_k |X(f_k)|
\end{equation}
Peak frequency highlights the dominant oscillatory behavior of the signal.

\emph{Maximum FFT Magnitude $\mathsf{max_f}$}:
The maximum magnitude of the FFT, excluding the DC component $(f=0)$, is given by:
    \begin{equation}
    \mathsf{max_f} =\max_k |X(f_k)|
    \end{equation}
This feature reflects the strongest frequency component in the signal.

The chosen statistical and spectral features were chosen for their proven effectiveness in characterizing vibration signals and detecting changes in structural response, as supported by \cite{buckley2023feature}. Variance and RMS quantify the energy and amplitude level of the vibration response, reflecting local modal participation and the degree of dynamic amplification experienced at the sensor location. Sensors exhibiting similar energy content often share comparable modal influence, making these metrics effective indicators of response similarity. Skewness and kurtosis describe the symmetry and peakedness of the vibration response, distinguishing sensors affected by nonlinear or boundary effects, such as those located near supports or joints, from those dominated by linear, near-Gaussian vibrations. These higher-order statistics thus capture response irregularities that are critical for identifying damage-sensitive locations.

In the frequency domain, spectral features such as the dominant frequency, peak magnitude, and energy in low-frequency bands capture modal characteristics and coupling effects, enabling the grouping of sensors that respond to the same structural modes. Together, these features provide a compact yet physically meaningful representation of structural behavior, linking measurable signal properties to the underlying dynamics.

The process of extracting statistical and spectral features from sensor signals is systematically summarized in Algorithm \ref{alg:feature_extraction} on the Appendix.

\subsubsection{Clustering via $K$-Means}
To select $M$ sensors, we first cluster the feature vectors $\{\mathbf{f}_1, \mathbf{f}_2, \ldots, \mathbf{f}_N\}$ into $M$ distinct groups, denoted as $C = \{C_1, C_2, \ldots, C_M\}$. Any clustering algorithm that  supports specifying the number of clusters can be used for this task such as $K$-Means and Gaussian Mixture Model. In this work, we adopt the $K$-Means clustering algorithm owing to its efficiency and scalability for large datasets. The objective of the clustering step  is to minimize the within-cluster variance, thereby ensuring that sensors grouped within the same cluster exhibit similar feature representations :

\begin{equation}
\arg \min_{C} \sum_{k=1}^M \sum_{i \in C_k} \|\mathbf{f}_i - \mathbf{\mu}_k\|^2,
\end{equation}
where $\mathbf{\mu}_k$ represents  the centroid of cluster $C_k$. This approach  ensures that sensors grouped within each cluster exhibit similar signal characteristics, thereby reducing redundancy while retaining the diversity of the original dataset.

\subsubsection{Graph-Based Representative Selection within Clusters}
To identify  the most representative sensor in each cluster, we incorporate graph-based centrality measures to evaluate the importance of each sensor within the network topology. These measures capture various aspects of a node’s structural and functional significance, ensuring a well-rounded selection process.

\emph{Degree Centrality} ($\text{deg}(v_i)$): Degree centrality measures the number of direct connections a sensor node \( v_i \) has in the network, representing its immediate connectivity. For a node \( v_i \) in a graph \( \mathcal{G} = (\mathcal{V}, \mathcal{E}, \mathbf{A}) \), it is defined as:

\begin{equation}
\text{deg}(v_i) = \sum_{j \in \mathcal{V}} \mathbf{A}(i, j),
\end{equation}
where \( \mathbf{A}(i,j) \) is the \( (i,j) \) entry of the adjacency matrix \( \mathbf{A} \) of graph \( \mathcal{G} \). High degree centrality indicates a sensor's role as a “hub” in the network with extensive local connectivity.

\emph{Betweenness Centrality} ($\text{betw}(v_i)$): Betweenness centrality measures the extent to which a sensor node \( v_i \) lies on the shortest paths between other nodes, highlighting  its role as a “bridge” in information flow. For a node \( v_i \), betweenness centrality is defined as:

\begin{equation}
\text{betw}(v_i) = \sum_{j \neq i \neq k} \frac{\sigma_{jk}(v_i)}{\sigma_{jk}},
\end{equation}
where \( \sigma_{jk} \) is the total number of shortest paths between nodes \( j \) and \( k \), and \( \sigma_{jk}(v_i) \) is the number of those shortest paths passing through node \( v_i \). Betweenness centrality highlights nodes that act as connectors within the network, bridging different parts of the graph. Sensors with high betweenness centrality serve as critical links in the data flow, often mediating information exchange between otherwise disconnected clusters. This is particularly valuable in IoT networks where certain sensors enable communication across distinct areas of the infrastructure.

\emph{Eigenvector Centrality} ($\text{eig}(v_i)$): Eigenvector centrality measures the influence of a node based on the importance of its neighbors, assigning higher centrality to nodes connected to other highly connected nodes. The eigenvector centrality for node \( v_i \) is defined as the \( i \)-th entry of the eigenvector \( \mathbf{e} \) corresponding to the largest eigenvalue \( \lambda_{\text{max}} \) of the adjacency matrix \( \mathbf{A} \):

\begin{equation}
\lambda_{\text{max}} \mathbf{e}(i) = \sum_{j \in \mathcal{V}} \mathbf{A}(i,j) \mathbf{e}(j),
\end{equation}
Eigenvector centrality considers not just the connectivity of a node but also the quality of its connections, assigning higher values to nodes that are linked to other influential nodes. In sensor networks, this centrality measure can help identify key nodes that have broader influence across the network, making them valuable for monitoring or control points within the system.

\emph{Harmonic Centrality} ($\text{harm}(v_i)$): Harmonic centrality captures the local efficiency of a node \( v_i \), calculated as the sum of the reciprocals of the shortest path distances from \( v_i \) to all other nodes \( v_j \):

\begin{equation}
\text{harm}(v_i) = \sum_{\substack{j \in \mathcal{V} \\ j \neq i}} \frac{1}{d(v_i, v_j)},
\end{equation}
where \( d(v_i, v_j) \) represents the shortest path distance between nodes \( v_i \) and \( v_j \). Harmonic centrality remains well-defined even for disconnected graphs and provides a measure of a node’s ability to efficiently reach other nodes in the network. This local emphasis is valuable for sensor networks where proximity-based interactions are critical, as it highlights sensors that can efficiently exchange information with nearby sensors.

\emph{Localization Operator} ($\text{loc}(v_i)$): Given a graph \( \mathcal{G} \) with a graph Laplacian matrix \( \mathbf{L} \), the localization operator \( \mathbf{T}_g \) is derived from the eigenstructure of \( \mathbf{L} \). The Laplacian's eigenvalue decomposition provides the eigenvalues \( \lambda_l \) and corresponding eigenvectors \( \mathbf{u}_l \). The localization operator for node \( i \) is defined by the expression \cite{perraudin2018global}:
\begin{equation}
\mathbf{T}_{g,i}(n) = \sqrt{N} \sum_{l=0}^{N-1} g(\lambda_l) \mathbf{u}_l(i) \mathbf{u}_l(n),
\end{equation}
where \( g(\lambda_l) \) is a filter kernel applied to the eigenvalue \( \lambda_l \), and \( \mathbf{u}_l(i) \) and \( \mathbf{u}_l(n) \) represent the \( i \)-th and \( n \)-th components of the eigenvector \( \mathbf{u}_l \), respectively. The matrix \( \mathbf{T}_g \) can be assembled as:
\begin{equation}
\mathbf{T}_g = \mathbf{U} g(\boldsymbol{\Lambda}) \mathbf{U}^\mathsf{T},
\end{equation}
where \( \mathbf{U} \) is the matrix of eigenvectors of the Laplacian \( \mathbf{L} \), and \( \boldsymbol{\Lambda} \) is the diagonal matrix of its eigenvalues. The feature vector for each node is extracted from the localization operator matrix \( \mathbf{T}_g \). To obtain a scalar value representing the node's spectral behavior, the $\ell_2$ norm (Euclidean norm) of each row of the localization matrix is computed. The \( i \)-th node’s localization feature vector is given by:
\begin{equation}
\text{loc}(v_i) =\| \mathbf{T}_{g,i} \|_2
\end{equation}
This $\ell_2$ norm of the \( i \)-th row of the matrix quantifies the spectral influence of node \( i \) across the graph, capturing both local and global structural properties. 

\begin{figure*}
    \centering
    \includegraphics[width=\linewidth]{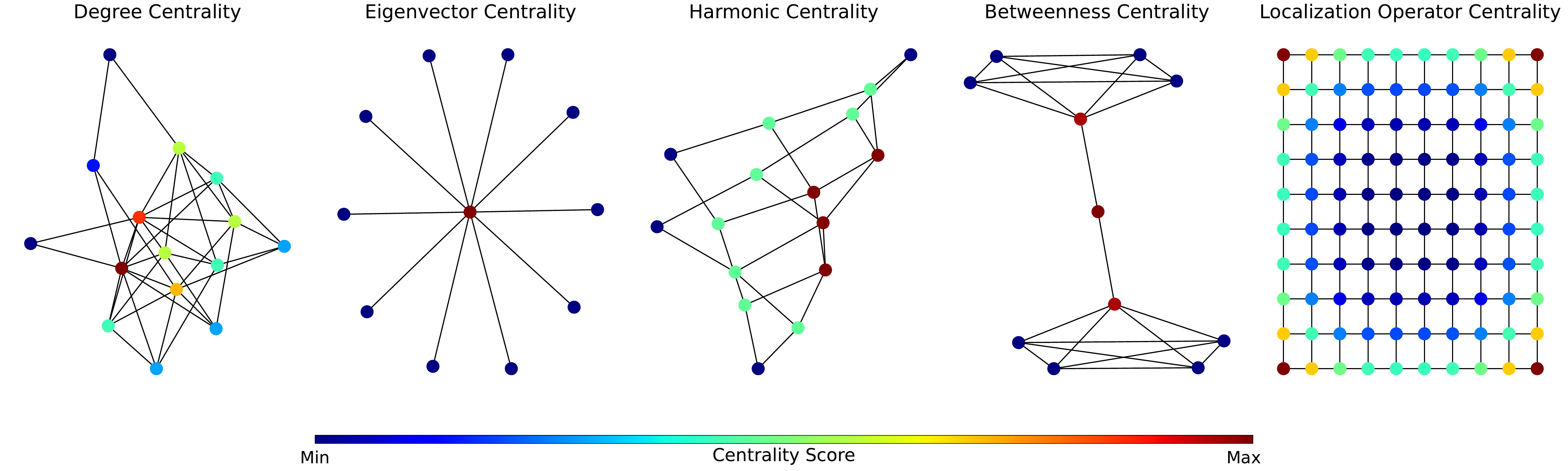}
    \caption{Visualization of Various Node Centrality Measures in Network Graphs. }
    \label{fig:Centralities}
\end{figure*}

Figure \ref{fig:Centralities} provides a detailed visualization of various node centrality measures within network graphs, illustrating the distinct ways to assess and characterize  nodes' importance and influence in the network topology. 
Algorithm \ref{alg:centrality_calculation} in the Appendix presents  a systematic  methodology for extracting structural node features, providing a comprehensive framework for analyzing and quantifying the roles of nodes in complex networks.

To determine  the most representative sensor within each cluster \( C_k \), we calculate  an aggregated centrality score by combining various centrality measures with designated weighting factors. Let \( \mathbf{d} = [d_i] \) represent the degree centrality vector, where  \( d_i = \text{deg}(v_i) \) corresponds to the degree centrality of sensor node \( v_i \). Similarly, define  \( \mathbf{b} = [b_i] \) as the betweenness centrality vector, with  \( b_i = \text{betw}(v_i) \); \( \mathbf{e} = [e_i] \) as the eigenvector centrality vector, where \( e_i = \text{eig}(v_i) \); \( \mathbf{h} = [h_i] \) as the harmonic centrality vector, with \( h_i = \text{harm}(v_i) \)  and \( \mathbf{l} = [l_i] \) as the localization feature vector, where \( l_i = \text{loc}(v_i) \). The relative importance of each centrality metric is encoded in a weighting  vector \( \boldsymbol{\alpha} = [\alpha_0, \alpha_1, \alpha_2, \alpha_3, \alpha_4]^\mathsf{T} \), where each component \(\alpha_k  \) specifies  the weight assigned to the corresponding centrality measure.  This framework ensures flexibility in tailoring the importance of different metrics to specific application requirements, thereby providing a robust evaluation of sensor representativeness.

The TopoScore \( \psi_i \) for each sensor node within a cluster is computed by weighting and summing its centrality measures as follows:

\begin{equation}
\psi_i = \alpha_0 \cdot d_i + \alpha_1 \cdot b_i + \alpha_2 \cdot e_i + \alpha_3 \cdot h_i + \alpha_4 \cdot l_i.
\label{topoeq}
\end{equation}
where \(d_i\), \(b_i\), \(e_i\), \(h_i\), and \(l_i\) denote the degree, betweenness, eigenvector, harmonic, and localization centrality values for node \(v_i\), respectively. The weights \(\alpha_0, \alpha_1, \alpha_2, \alpha_3, \alpha_4\) are user-defined parameters reflecting the relative importance of each centrality measure.

In matrix-vector notation, the TopoScore \( \boldsymbol{\psi} \) for all nodes in a cluster can be expressed compactly as:

\begin{equation}
\boldsymbol{\psi} = \mathbf{C} \cdot \boldsymbol{\alpha},
\end{equation}
where \( \mathbf{C} = \begin{bmatrix} \mathbf{d} & \mathbf{b} & \mathbf{e} & \mathbf{h} & \mathbf{l} \end{bmatrix} \) is a matrix comprising the centrality vectors as columns, each representing a different centrality measure and \( \boldsymbol{\alpha} = [\alpha_0, \alpha_1, \alpha_2, \alpha_3, \alpha_4]^\mathsf{T} \) is the weighting vector. 
The weighting vector $\boldsymbol{\alpha}$ could in principle be tuned using validation data or domain knowledge to emphasize the most relevant centrality measures. In our study, we adopt a uniform weighting, which ensures balanced contributions from all five measures and avoids bias toward any single criterion. Extreme distributions (e.g., $[0.8, 0.05, 0.05, 0.05, 0.05]$) would overemphasize one structural attribute, such as degree or eigenvector centrality, potentially neglecting complementary aspects of connectivity. Since our network contains no isolated nodes and no single node dominates the topology, applying uniform weights ensures that all relevant structural characteristics are fairly represented, making this approach principled for selecting a balanced set of sensors. Further quantitative justification for this choice, including a comprehensive sensitivity analysis over the weighting vector $\boldsymbol{\alpha}$, is presented in Appendix F.

To identify the most representative sensor in each cluster \( C_k \), we select the node \( v_{i^*} \) with the highest TopoScore:

\begin{equation}
i^* = \arg \max_{i \in C_k} \left( \psi_i \right).
\end{equation}

The final  selected set \( \mathcal{S}^* = \{ v_{i^*} \mid i^* \in C_k, \, k = 1, 2, \ldots, M \} \) represents the \( M \) most informative sensors based on a balanced consideration of connectivity, bridging role, influence, and local accessibility within the network.

\subsection{Spatial-Temporal Graph Neural Networks for Downstream Tasks}
To evaluate the effectiveness of the proposed sensor placement method, particularly in retaining the overall information from the original dataset, we assess its impact on two key downstream tasks, damage detection and time-varying (temporal) graph signal reconstruction. Damage detection evaluates whether the  subset of selected sensors provides  sufficient information to  accurately identify structural damage. It serves as a critical benchmark for determining the practical utility of the selected sensors in real-world monitoring scenarios. Graph signal reconstruction focuses on estimating  sensor values at unselected locations based on the measurements from the selected sensors, enabling effective structural health monitoring and predictive maintenance with fewer physical sensors \cite{niresi2024physics}. It measures the capability of the selected subset to represent the entire sensor network effectively. These tasks serve as benchmarks to demonstrate the efficacy of the selected sensors for SHM systems. To model spatial-temporal dependencies effectively, we employ  an STGNN framework. The adopted STGNN architecture utilizes  an encoder-decoder structure to effectively process  input tensors representing time-varying, graph-structured data. Within the encoder, the data undergoes   three  sequential processing stages: (1) diffusion graph convolution, which captures spatial dependencies by modeling the flow of information across the graph nodes, (2) convolution, which extracts local temporal features, highlighting short-term patterns in the data, and (3) a long short-term memory (LSTM) layer, which models long-range temporal dependencies, enabling the network to learn from sequential patterns over time. 

First, the diffusion graph convolution layer models spatial dependencies in the graph by performing a diffusion process over the graph structure. Mathematically, the operation is defined as \cite{li2018diffusion}:
\begin{equation}
\mathbf{H} = \sum_{k=0}^{K} \phi\left(\mathbf{P}^k \mathbf{X} \mathbf{W}^{(k)}\right),
\end{equation}
where \(\mathbf{P} = \mathbf{D}^{-1}\mathbf{A}\) is the normalized transition matrix, \(\mathbf{X}\) represents the node features, $\phi$ is the activation function, and \(k\) is the diffusion step. This process captures multi-hop neighborhood information for each node, enriching the spatial feature representation. The resulting representations are then passed to convolution layers, which operate along the time window dimension to extract temporal features. Next, the new representations are passed to an LSTM layer to model long-term temporal dependencies. The LSTM outputs a temporally and spatially enriched latent representation.

\begin{figure*}
    \centering
    \includegraphics[width=0.9\linewidth]{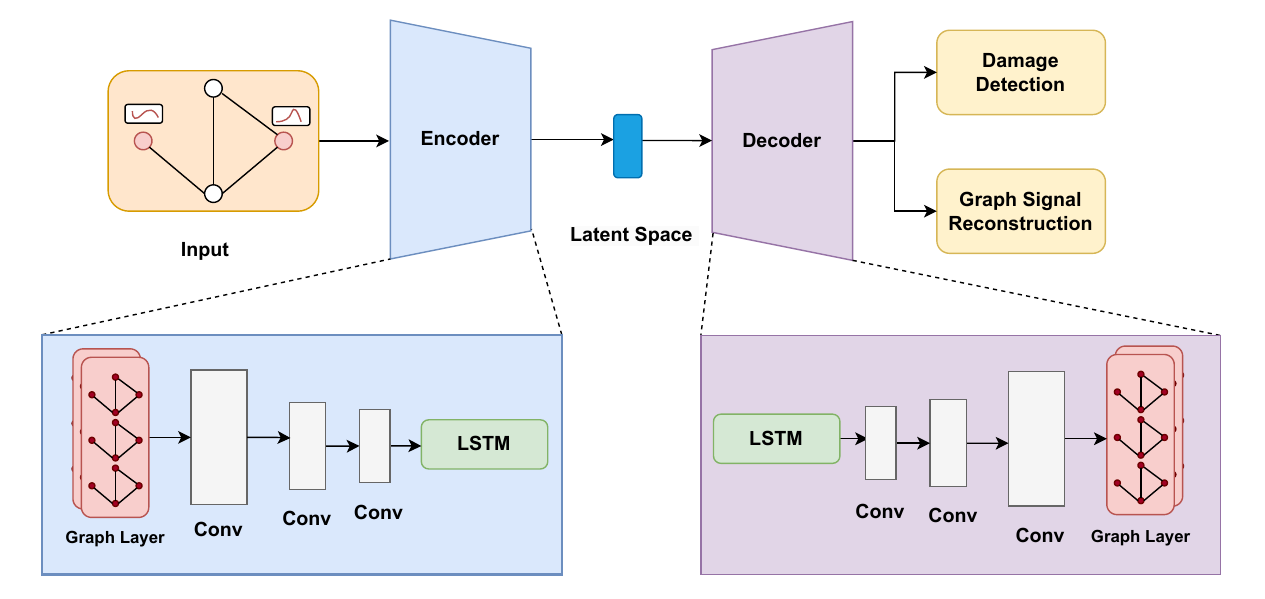}
    \caption{STGNN Architecture for Evaluation on Damage Detection and Graph Signal Reconstruction Task}
    \label{fig:stgnn}
\end{figure*}

In the decoder, the process mirrors the encoder in reverse. The decoder begins with an LSTM layer, which reconstructs the temporal patterns of the latent representation back into a sequence. The LSTM outputs are then passed to convolution layers, which refine the reconstructed sequence by applying learnable transformations along the time dimension, ensuring temporal coherence. Finally, the diffusion graph convolution layer reconstructs the spatial dependencies in the graph signals using the same diffusion process as the encoder. This ensures that the reconstructed signals respect the original graph structure and spatial relationships. Together, the encoder-decoder pipeline effectively captures and reconstructs the spatial-temporal dynamics of the graph-structured data. Moreover, to ensure fair comparison across all sensor selection strategies, the same graph topology representing the spatial relationships among sensors was used for every method. STGNN was applied using this fixed graph structure, while only the nodes corresponding to the selected sensors were retained in the model input. Unselected nodes were masked during feature propagation, allowing each method to be evaluated under identical network and training conditions. Figure \ref{fig:stgnn} illustrates the overview of STGNN architecture used in our work.

\subsection{Distinction from Existing Methods}
Conventional methods for vertex-domain sampling of graph signals often rely on predefined models such as the bandlimited model, the approximately bandlimited model, and the piecewise-smooth model. While these models provide a theoretical foundation for analytical sampling, a major limitation of these approaches is the potential mismatch between the assumed graph signal model and the actual underlying model in a given task. In practical scenarios, the true underlying graph signal model may deviate significantly from these assumptions, leading to significant performance degradation in real-world applications.
To overcome this challenge, we propose a time-vertex machine learning framework that employs a data-driven approach to model graph signals. Unlike traditional methods that depend on predefined assumptions, this framework offers the flexibility to adapt to diverse graph signal structures embedded in the data, making it more robust and applicable across varying scenarios. While several components of the proposed framework, such as feature extraction, clustering, and graph-based analysis, 
are individually established in the literature, the core novelty of this work lies in the unified modeling of the time–vertex domain 
for interpretable and efficient sensor placement in SHM. The TVML framework introduces a joint time–vertex modeling paradigm that simultaneously captures temporal behavior via statistical and spectral features, and spatial relationships through graph centrality measures, enabling a holistic representation of sensor dynamics fused by the TopoScore metric to identify representative sensors. It is an interpretable and model-agnostic framework that, unlike assumption-heavy existing approaches, remains data-driven to learn patterns directly from measurements while maintaining interpretability through physically meaningful features and network descriptors.

Similarly, graph signal recovery techniques traditionally solve optimization problems using predefined graph-regularization terms, such as the quadratic form of the graph Laplacian. However, a significant challenge lies in selecting an appropriate regularization term for a specific recovery task. An improperly chosen regularization can introduce bias, resulting in suboptimal performance in real-world applications. In this work, we leverage STGNNs to incorporate inductive biases, effectively encoding prior knowledge about the graph signals directly within the model, rather than relying on predefined regularization terms.

\section{Experimental Results}
\label{sec:experiments}

\subsection{Case Studies}
To assess our method, we selected two different case studies that demonstrate the challenges and techniques associated with bridge health monitoring (BHM). The first case study is about the Train-Track-Bridge dataset, developed using a numerical model of a coupled train-track-bridge system \cite{sarwar2024probabilistic}, which makes it possible to study the influence of structural damage on the dynamic responses of the structure, allowing us to perform damage detection. The second case study examines the Long-span Cable-stayed Bridge dataset which provides large amounts of acceleration measurements collected from a long-span bridge \cite{bao20211st}. This allows for the analysis of high-frequency data reconstruction in the presence of numerous sensor failures.

\subsubsection{Train-Track-Bridge Dataset} The Train-Track-Bridge dataset is generated through a numerical model that simulates the interactions between a train, ballasted track, and the bridge, as illustrated in Figure \ref{fig:graph_and_heatmap}. We created ten sensor locations capable of measuring displacement and acceleration. In the numerical simulation, structural damage was introduced by reducing the stiffness of the bridge elements near the mid-span to emulate localized stiffness degradation. The bridge was discretized into 160 finite elements, each approximately 0.3 m in length, and damage was modeled by reducing the stiffness of two adjacent elements, corresponding to about 1.2\% of the total bridge length. This reduction represents physical deterioration mechanisms such as cracking or reinforcement debonding, which decrease bending rigidity while preserving overall geometry and mass distribution. 
The stiffness-reduction approach, commonly adopted in train–track–bridge studies \cite{sarwar2024probabilistic}, allows controlled variation of damage severity and provides a realistic means to study its influence on dynamic responses. Additionally, labels indicating the condition of the bridge—either healthy or damaged—were generated, with varying degrees of damage represented. Approximately 400 samples were produced, each corresponding to the measurements taken as a train traverses the bridge track. Each sample includes a label indicating the bridge's state, along with the recorded displacement and acceleration signals from the ten designated sensor locations during the passage of the train. The dataset contains 8.24 million labeled timesteps in total: 3.84M corresponding to healthy conditions and 4.40M to damaged conditions. These timesteps are grouped into sets used for training, validation, and testing. The training set includes healthy data (20 segments of 20,000 timesteps each). The validation set includes both healthy and damaged data (20 segments each). The test set includes 152 healthy and 200 damaged segments. While the training set uses a limited number of segments, each segment is long (20,000 timesteps), ensuring that the dataset overall is large and information-rich.

\subsubsection{Long-span Cable-stayed Bridge Dataset} The dataset is collected from a long-span cable-stayed bridge in China, and the measurements were taken under harsh environmental conditions \cite{bao20211st}. 
The dataset utilized in this study comprises acceleration measurements continuously recorded over a one-month period (January 1–31, 2012) at a sampling frequency of 20 Hz.  A total of 38 accelerometers were strategically installed on the deck and towers of the bridge to capture both vertical and horizontal dynamic responses. This dense spatial distribution ensures a comprehensive depiction of the bridge’s structural behavior under varying environmental and operational conditions. This dataset includes six types of abnormal cases affecting the sensors, which include outliers, minor deviations, drift, trends, missing values, and square waves, including mixtures of these abnormalities. Consequently, for our analysis, we selected one hour of data from the sensors oriented in the X direction, excluding those exhibiting anomalies and focusing solely on the normal sensors. Specifically, we selected ten sensors, namely, 4, 6, 8, 10, 12, 26, 29, 31, 33, and 36 for analysis. 

\subsection{Data Preprocessing and Implementation Details}

The data preprocessing pipeline was designed to prepare sequential sensor data for downstream tasks, such as damage detection and time-varying graph signal reconstruction. The input consists of a time-series signal from multiple sensors. The goal is to reconstruct graph signal values for nodes within the graph, enabling accurate time-varying graph signal reconstruction and damage detection. For both tasks, the training data exclusively consists of healthy samples. This approach ensures that the model learns the underlying patterns of normal behavior, enabling effective anomaly detection by identifying deviations and supporting accurate reconstruction by leveraging the inherent structure of healthy graph signals. Moreover, the dataset is split chronologically to maintain the temporal order of events, with earlier time periods used for training and later periods reserved for validation and testing.

Before feeding data into the model, normalization is applied based on the statistics of the training dataset to standardize feature values. This normalization is computed as:
\begin{equation}
    \mathbf{X}_{\text{norm}} = \frac{\mathbf{X} - \boldsymbol{\mu}_{\text{train}}}{\boldsymbol{\sigma}_{\text{train}}}
\end{equation}
where \(\mathbf{X}\) is the original signal value, \(\boldsymbol{\mu}_{\text{train}}\) and \(\boldsymbol{\sigma}_{\text{train}}\) are the mean and standard deviation of the training set, respectively, and \(\mathbf{X}_{\text{norm}}\) is the normalized value. To prepare the data for time-series analysis, a sliding window approach is employed. A batch size of 64 is used for training, and the Adam optimizer is applied with a learning rate of 0.003. To prevent overfitting, an early stopping mechanism is implemented, which stops training if the validation loss does not improve over 30 consecutive epochs. All other hyperparameters are summarized in the Appendix.

\subsection{Evaluation Metrics}

To assess the performance of the proposed framework, we utilize different evaluation metrics tailored to the specific tasks of damage detection and signal reconstruction. For the damage detection task, we employ precision, recall, accuracy, Receiver Operating Characteristic (ROC), Area under the Curve (AUC), and F1 score, which quantify the model’s fault detection performance.  For the signal reconstruction task, we use the Root Mean Square Error (RMSE) and the Mean Absolute Error (MAE) to evaluate reconstruction accuracy in a scale-independent manner. Detailed mathematical formulations of these metrics are provided in the Appendix.

\subsection{Results and Discussion}
\subsubsection{Damage Detection}

\begin{figure}[H]
    \centering
    \begin{subfigure}[b]{0.9\textwidth}
        \centering
        \includegraphics[width=\textwidth]{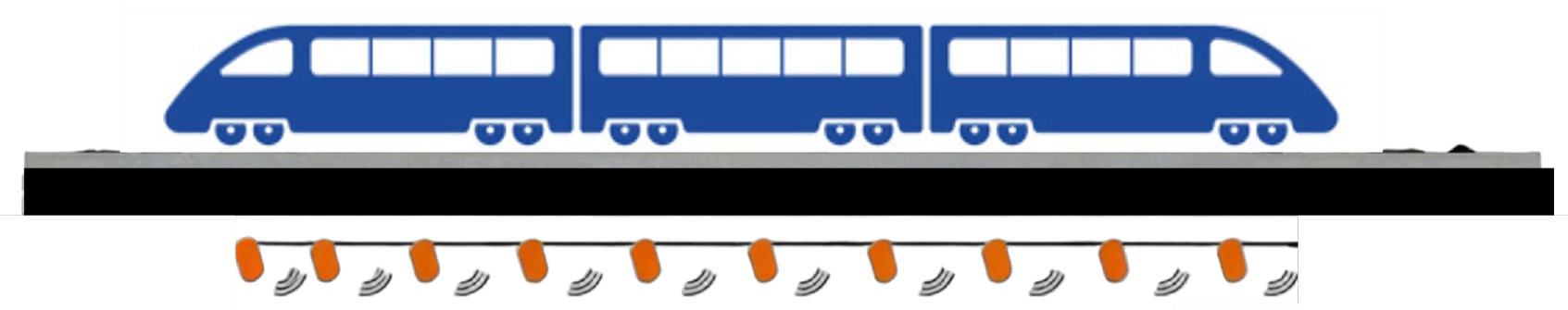} 
        \caption{Train-Track-Bridge Physical Structure.}
        \label{fig:overview_structure1}
    \end{subfigure}
    
    \vspace{0.5cm} 

    \begin{subfigure}[b]{0.55\textwidth}
        \centering
        \includegraphics[width=\textwidth]{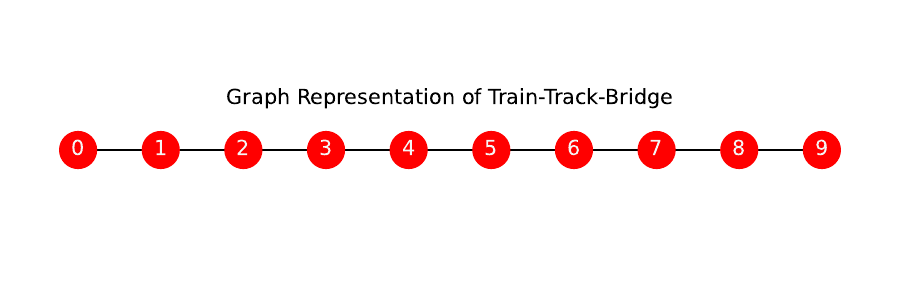}
        \caption{Graph Representation of Train-Track-Bridge.}
        \label{fig:graph_representation}
    \end{subfigure}
    \begin{subfigure}[b]{0.35\textwidth}
        \centering
        \includegraphics[width=\textwidth]{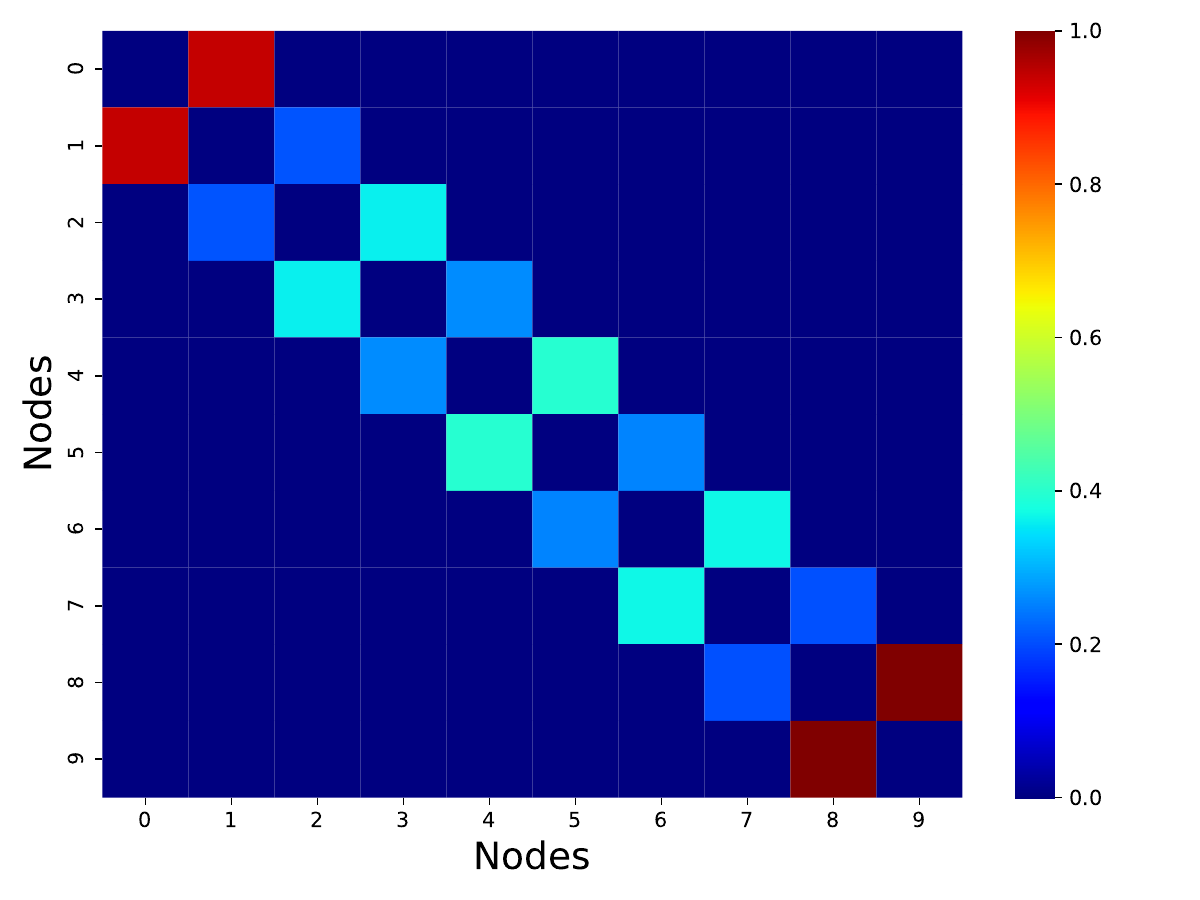}
        \caption{Colormap of Adjacency Matrix.}
        \label{fig:adjacency_heatmap1}
    \end{subfigure}

    \caption{Comprehensive visualization of the Train-Track-Bridge dataset. (a) Train-Track-Bridge physical structure. (b) The graph representation aligns all nodes in a straight line, resembling the physical structure of the bridge. (c) The colormap of the adjacency matrix visualizes connectivity strengths among nodes.}
    \label{fig:graph_and_heatmap}
\end{figure}

For the damage detection task, we utilized the Train-Track-Bridge dataset. In the first step, we learned the graph structure based on a smooth graph signal representation as defined in (\ref{eq:kalo}). The resulting graph, depicted in Figure \ref{fig:graph_and_heatmap}, consists of nodes representing sensors and weighted edges that indicate the connectivity between these sensors. As a post-processing step, we scaled the adjacency matrix of the graph. The resulting structure closely resembles the physical configuration of the bridge, forming a line graph. An interesting observation from this learned graph is that the edge weights are not uniform across the sensors, despite the sensors being evenly distributed along the bridge. This variation in edge weights can be attributed to the complex structural characteristics of the beam bridge. Factors such as structural geometry, dynamic load distribution, and support conditions contribute to these disparities, reflecting varying degrees of interaction and connectivity between different sections of the bridge.

These observations highlight the value of the graph learning approach, which inherently accounts for the structural complexity of the bridge. By capturing these interactions, the learned graph provides a more accurate representation of the underlying dynamics, enhancing the analysis and interpretation of sensor data for effective damage detection.

We perform anomaly detection using an autoencoder trained solely on healthy data. The autoencoder is trained to reconstruct input samples \( \mathbf{X} \in \mathbb{R}^{B \times T \times N \times F} \), where \( B \) is the batch size, \( T \) is the temporal length, \( N \) is the number of nodes (e.g., locations), and \( F \) is the feature dimension. The reconstruction loss is minimized, and the model learns to capture the healthy patterns of the system.  For anomaly detection, the RMSE is computed for each sample. The detection threshold 
$\delta$ is defined based on healthy part of validation data as $ \delta = \mu_{\text{healthy}} + 2 \sigma_{\text{healthy}}$, where $\mu_{\text{healthy}}$ and $\sigma_{\text{healthy}}$ denote the mean and standard deviation of RMSE values from the healthy part of validation set, respectively. Test samples with RMSE exceeding this threshold are classified as anomalous.

For the experiments, displacement sensors were chosen as the primary sensor type because they provided the most effective separation between healthy and damaged states. Specifically, this sensor type exhibited the largest divergence in reconstruction error between normal and anomalous conditions based on validation set, making it particularly suitable for structural damage detection in our study. Finally, we compared the proposed sensor selection method (TVML) to other approaches (Entropy \cite{krause2008near, shewry1987maximum}, MI \cite{sharma2015greedy, krause2008near}, Random \cite{puy2018random}, QR Pivoting \cite{manohar2018data}, and Localization operator \cite{sakiyama2019eigendecomposition}) with the same sampling ratio of 0.2.

Table \ref{tab:results_dd} presents a comparative evaluation of six methods for the damage detection task, assessed using five key performance metrics: Accuracy, Precision, Recall, F1 Score, and AUC. Collectively, these indicators provide a balanced perspective on each method’s capability to identify structural damage while managing false positives and false negatives. Among the baseline methods, the Random and Localization approaches yield the most competitive results, with F1 scores of 0.813 and 0.802, respectively, indicating consistent detection performance. The QR Pivoting method also performs reasonably well (F1 = 0.772), while the Entropy and MI methods show moderate detection capabilities, achieving F1 scores of 0.746 and 0.765, respectively. Notably, the proposed TVML framework outperforms all baselines across every metric, achieving the highest accuracy (0.864), F1 score (0.886), and AUC (0.916). Its superior recall (0.935) demonstrates exceptional sensitivity to damage, while maintaining strong precision (0.842). These results confirm TVML’s robustness and effectiveness in capturing the complex spatial-temporal dynamics of structural responses, leading to more reliable and consistent damage detection than existing methods.

\begin{table}[ht]
\centering
\caption{Comparison of Methods for Damage Detection Task}
\label{tab:results_dd}
\begin{tabular}{lccccc}
\toprule
\textbf{Method} & \textbf{Accuracy} & \textbf{Precision} & \textbf{Recall} & \textbf{F1 Score} & \textbf{AUC} \\ 
\midrule
Random        & 0.787 & 0.811 & 0.815 & 0.813 & 0.829 \\
Localization  & 0.773 & 0.794 & 0.810 & 0.802 & 0.816 \\
Entropy       & 0.716 & 0.758 & 0.735 & 0.746 & 0.762 \\
MI            & 0.733 & 0.765 & 0.765 & 0.765 & 0.794 \\
QR Pivoting   & 0.750 & 0.801 & 0.745 & 0.772 & 0.781 \\
\textbf{TVML} & \textbf{0.864} & \textbf{0.842} & \textbf{0.935} & \textbf{0.886} & \textbf{0.916} \\ 
\bottomrule
\end{tabular}
\end{table}

\begin{figure}[H]
    \centering
    \includegraphics[width=0.75\linewidth]{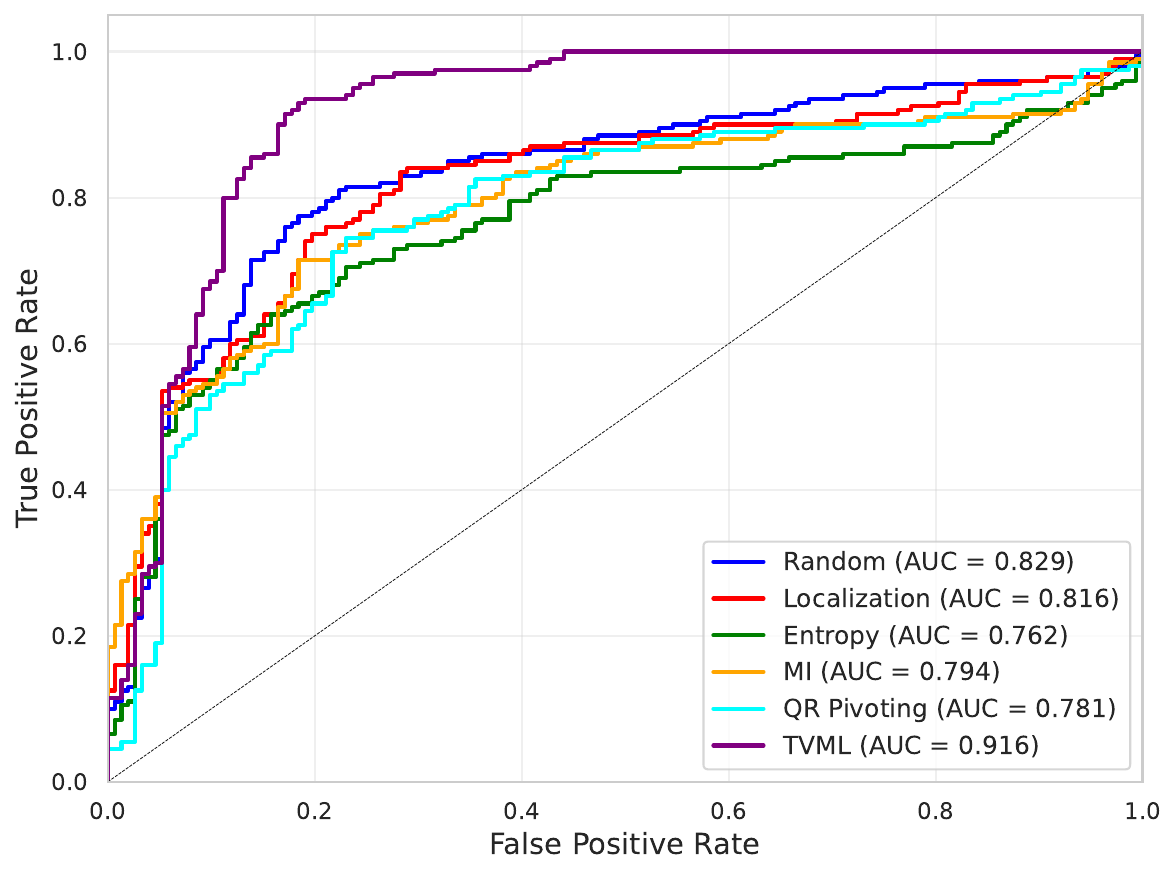}
    \caption{ROC curves for the compared methods}
    \label{fig:roc}
\end{figure}

The performance of different methods was comprehensively analyzed using a variety of visualizations to highlight their strengths and weaknesses in terms of classification metrics. Figure \ref{fig:roc} presents the ROC curves that compare the performance of various methods in the damage detection task. The plot shows the True Positive Rate (TPR) on the y-axis and the False Positive Rate (FPR) on the x-axis, with each curve covering the full range of classification thresholds. The diagonal dashed line represents the baseline performance, corresponding to random guessing, and serves as a reference for evaluating model effectiveness. The TVML curve outperforms the others, with its trajectory deviating upwards from the random baseline, indicating superior discriminatory power. Notably, the TVML curve approaches the ideal top-left corner, demonstrating high sensitivity and specificity, while other methods show more moderate performance. These results underscore the effectiveness of TVML in selecting the most discriminative sensors, effectively minimizing false positives while maximizing true positives.

\begin{figure}[h!]
    \centering
    \includegraphics[width=\textwidth]{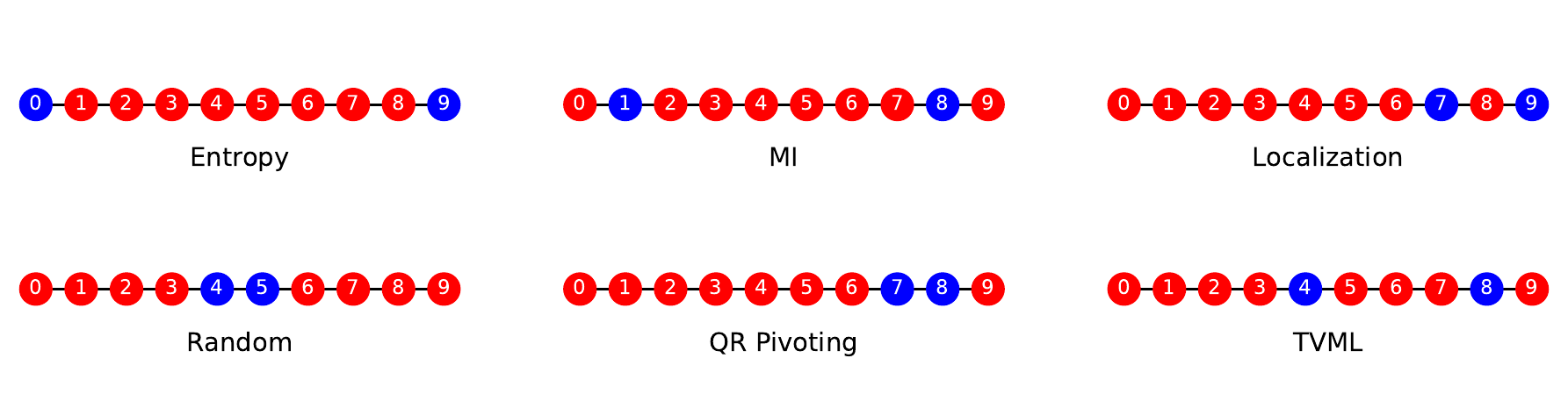}
    \caption{Visualization of selected sensors for the first case in each method. Blue nodes indicate selected sensors, and red nodes indicate unselected sensors. For Random, the two most frequently selected nodes over 50 independent runs are highlighted.}
    \label{fig:selected_sensors}
\end{figure}

To provide an intuitive comparison among sensor selection strategies, Figure \ref{fig:selected_sensors} illustrates the sensors chosen by each of the six methods for the first case study. Blue nodes denote selected sensors, and red nodes denote unselected sensors. It can be observed that the Entropy-based method tends to select sensors located at the boundaries of the graph, as these nodes typically have fewer connections with other nodes. In contrast, the other baseline methods generally produce more evenly distributed sensor locations across the structure.

\textbf{Sensitivity Analysis on the Number of Selected Sensors:} To provide a transparent understanding of how the desired number of selected sensors \(M\) influences performance, 
we conducted a quantitative sensitivity analysis by varying \(M\) from 10\% to 80\% of the available sensors. For each configuration, the TVML framework was applied to perform the damage detection task using the same experimental settings described earlier.

\begin{figure}[h!]
    \centering
    \includegraphics[width=0.7\textwidth]{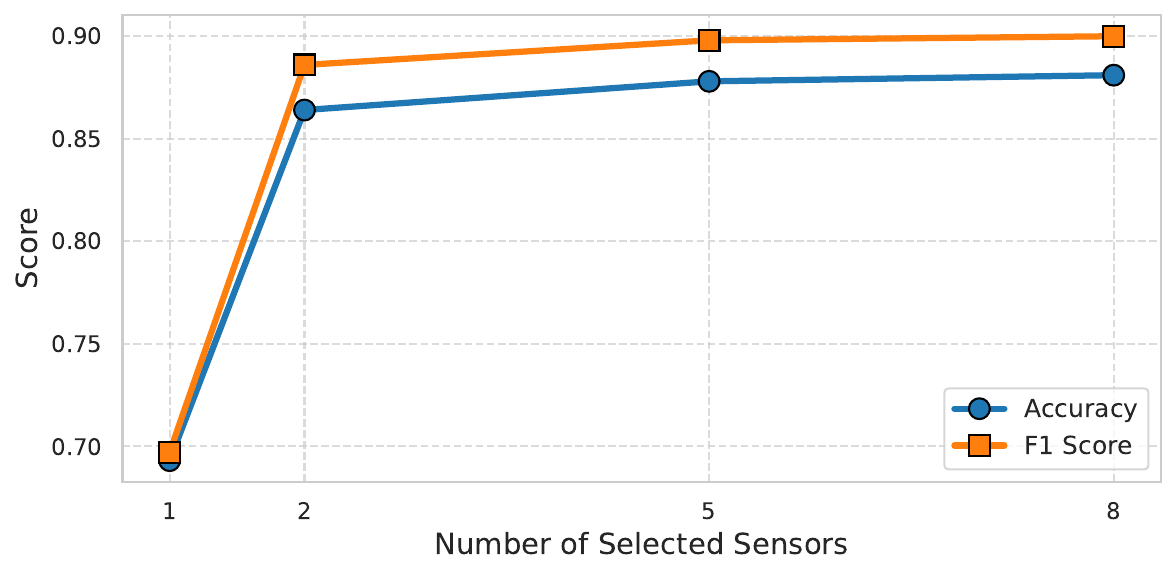}
    \caption{Sensitivity analysis of the number of selected sensors. 
    Damage detection F1-score for varying percentages of selected sensors (10\%–80\%), 
    showing substantial improvement up to approximately 20\% and stable performance thereafter.
    These results highlight the practical trade-off between sensor count and detection accuracy.}
    \label{fig:maskoptimization}
\end{figure}

The results, illustrated in Figure~\ref{fig:maskoptimization}, show that both the F1-score and Accuracy improve 
rapidly as the proportion of selected sensors increases from 10\% to 20\%.
Beyond this threshold, performance saturates, indicating diminishing returns for larger sensor deployments. Consequently, the desired sensor count \(M\) can be determined adaptively by defining acceptable performance thresholds 
for  damage detection accuracy. 
Specifically, one may select the smallest \(M\) such that $\text{Accuracy} > \zeta$, where \(\zeta\) represents user-defined tolerances for damage detection performance, respectively.

\subsubsection{Time-Varying Graph Signal Reconstruction}

\begin{figure}
    \centering
    \begin{subfigure}[b]{\textwidth}
        \centering
        \includegraphics[width=\textwidth]{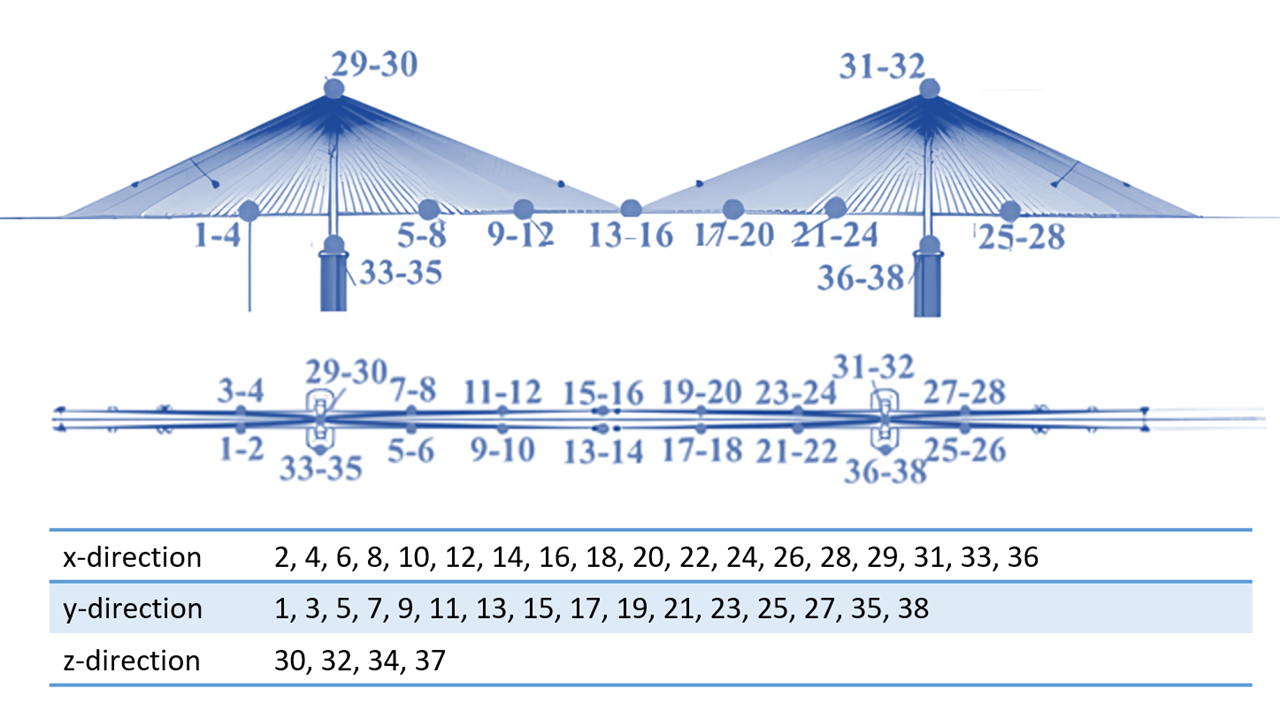} 
        \caption{Long-span Cable-stayed Bridge Physical Structure (adapted from \cite{bao20211st}).}
        \label{fig:overview_structure}
    \end{subfigure}
    
    \vspace{0.5cm} 

    \begin{subfigure}[b]{0.55\textwidth}
        \centering
        \includegraphics[width=\textwidth]{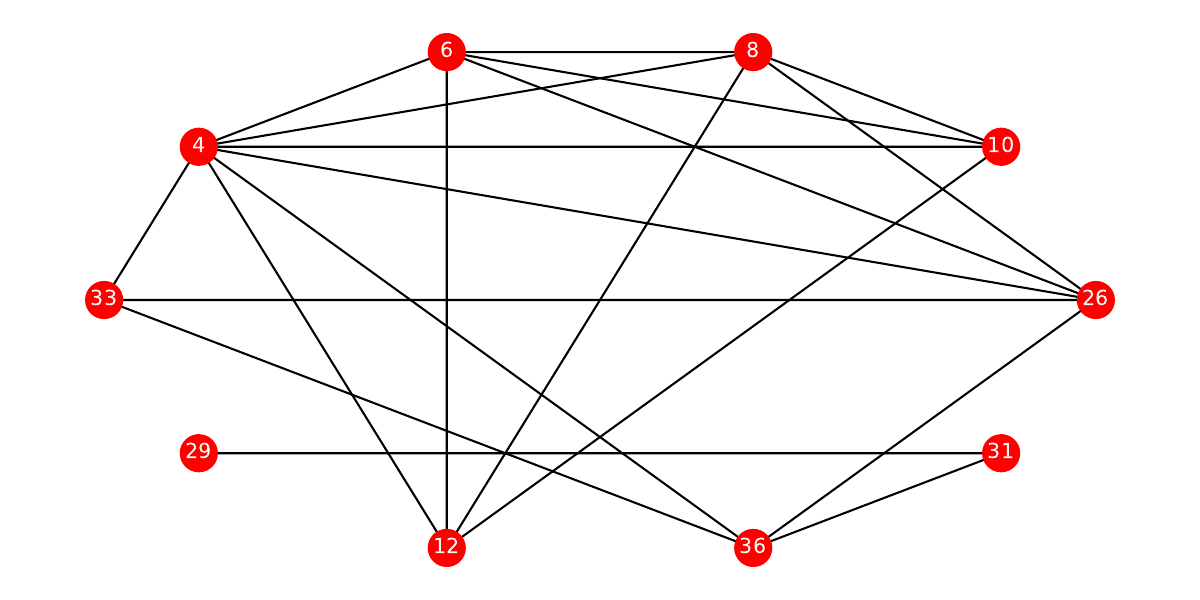}
        \caption{Graph Representation of Long-span Cable-stayed Bridge Dataset.}
        \label{fig:graph_representation_iPC}
    \end{subfigure}
    \begin{subfigure}[b]{0.35\textwidth}
        \centering
        \includegraphics[width=\textwidth]{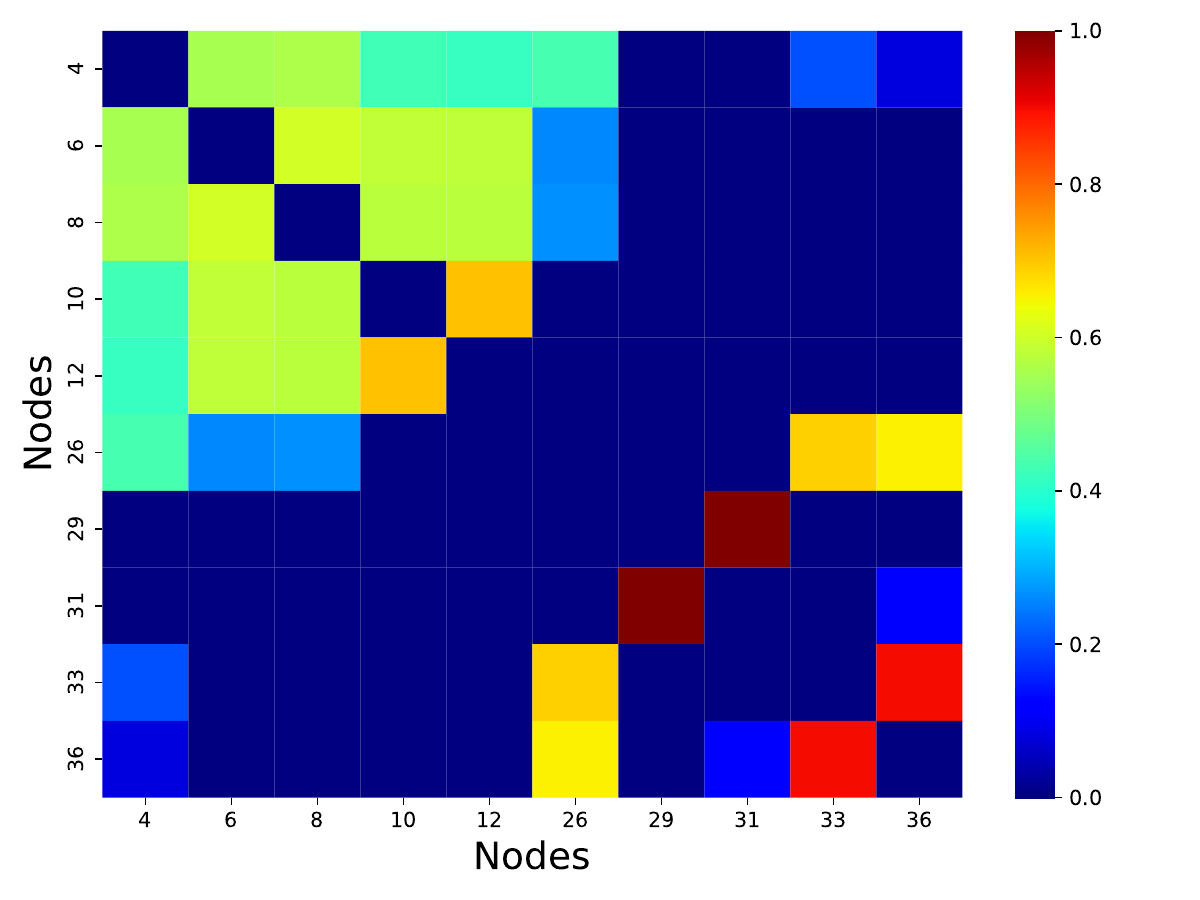}
        \caption{Colormap of Adjacency Matrix.}
        \label{fig:adjacency_heatmap}
    \end{subfigure}

    \caption{Comprehensive visualization of the Long-span Cable-stayed Bridge dataset. (a) Bridge physical structure. (b) The graph representation of Long-span Cable-stayed Bridge dataset. (c) The colormap of the adjacency matrix visualizes connectivity strengths among nodes.}
    \label{fig:graph_and_heatmap_IPC}
\end{figure}

Unlike the Train-Track-Bridge dataset, the Long-span Cable-stayed Bridge dataset contains only acceleration signals. As such, we construct the graph using these acceleration values. The resulting graph, shown in Figure \ref{fig:graph_and_heatmap_IPC}, represents nodes as sensors, with weighted edges indicating the connectivity between these sensors.

The results presented in Table~\ref{tab:recres} and Figure~\ref{fig:gsr} provide a comprehensive comparison of various methods for time-varying graph signal reconstruction, focusing on two key performance metrics: RMSE and MAE. The table summarizes reconstruction error across five sensor configurations, denoted as \( |\mathcal{S}^*| \), ranging from 2 to 8 sensors. Each method (Entropy, MI, Localization, Random, QR Pivoting, and TVML) was evaluated under identical conditions to assess reconstruction quality.

Across all cases, TVML consistently demonstrates superior performance, achieving the lowest RMSE and MAE values in nearly every configuration. For example, at \( |\mathcal{S}^*| = 2 \), TVML attains an RMSE of 0.8128, outperforming the next best method, Localization, at 0.8308. This advantage persists as the number of sensors increases, with TVML maintaining the lowest RMSE values across all cases, reaching 0.6299 at \( |\mathcal{S}^*| = 8 \).

A similar trend is observed for the MAE metric, where TVML again yields the smallest errors across all sensor configurations. For instance, at \( |\mathcal{S}^*| = 5 \) and \( |\mathcal{S}^*| = 7 \), TVML achieves MAE values of 0.4441 and 0.4327, respectively, indicating more precise signal recovery compared to the competing approaches. The improvements are especially notable in scenarios with fewer sensors, highlighting TVML’s effectiveness in capturing the underlying spatio-temporal dependencies even under sparse sampling conditions. Overall, these results confirm that TVML provides the most accurate and robust reconstruction performance among all evaluated methods.

\begin{table}[ht]
\centering
\caption{Comparison of Different Methods in Time-Varying Graph Signal Reconstruction}
\resizebox{\linewidth}{!}{
\begin{tabular}{cccccccc}
\toprule
\textbf{Case} & \textbf{Metric} & \textbf{Entropy} & \textbf{MI} & \textbf{Localization} & \textbf{Random} & \textbf{QR Pivoting} & \textbf{TVML} \\ 
\midrule
\multirow{2}{*}{$|\mathcal{S}^*| = 2$} & RMSE & 0.8560 & 0.8362 & \underline{0.8308} & 0.8311 & 0.8498 & \textbf{0.8128} \\ 
 & MAE & 0.6173 & 0.5941 & \underline{0.5875} & \underline{0.5875} & 0.6074 & \textbf{0.5813} \\ 
\midrule
\multirow{2}{*}{$|\mathcal{S}^*| = 3$} & RMSE & 0.8059 & 0.7867 & 0.7721 & \underline{0.7441} & 0.7574 & \textbf{0.7344} \\ 
 & MAE & 0.5438 & 0.5241 & 0.5137 & \underline{0.4965} & 0.5030 & \textbf{0.4946} \\ 
\midrule
\multirow{2}{*}{$|\mathcal{S}^*| = 5$} & RMSE & 0.7120 & 0.6794 & 0.6869 & 0.6987 & \underline{0.6724} & \textbf{0.6642} \\ 
 & MAE & 0.4706 & 0.4790 & 0.4738 & 0.4841 & \underline{0.4645} & \textbf{0.4441} \\ 
\midrule
\multirow{2}{*}{$|\mathcal{S}^*| = 7$} & RMSE & 0.6550 & 0.6615 & 0.6631 & 0.6682 & \underline{0.6542} & \textbf{0.6517} \\ 
 & MAE & 0.4544 & 0.4684 & 0.4685 & 0.4704 & \underline{0.4524} & \textbf{0.4327} \\ 
\midrule
\multirow{2}{*}{$|\mathcal{S}^*| = 8$} & RMSE & 0.6547 & 0.6578 & \textbf{0.6153} & 0.6497 & 0.6521 & \underline{0.6299} \\ 
 & MAE & 0.4442 & 0.4478 & \underline{0.4328} & 0.4641 & 0.4507 & \textbf{0.4312} \\ 
\bottomrule
\end{tabular}
}
\label{tab:recres}
\end{table}

In summary, the analysis clearly suggests that TVML provides the best performance in time-varying graph signal reconstruction in terms of both RMSE and MAE, particularly when the cardinality of the sensor set is smaller.

\section{Conclusion}
\label{sec:conclusion}
In this work, we proposed TVML to address sensor selection problems for time-varying graph signals. Our approach addresses an important research gap in current sensor selection methods by considering both spatial and temporal aspects of bridge behavior, making it more adaptive and responsive to dynamic changes in structural conditions. The proposed method is evaluated on two datasets and two downstream assessment tasks: graph signal reconstruction and damage detection. Our results demonstrated that the proposed method outperforms existing techniques, achieving state-of-the-art results. An interesting direction for future work is the exploration of unsupervised geometric deep learning methods, such as spatial-temporal graph autoencoders, for deep feature extraction. Leveraging these methods could enable the replacement of handcrafted features with automatically learned spatial-temporal representations, which may enhance the model's ability to capture complex patterns in dynamic systems. Another limitation is the use of fixed, user-defined weights for combining the sensor selection criteria. In our current setting, this design choice is reasonable because the network is relatively simple and does not contain nodes with highly unique behaviors or extreme centrality, so equal weighting ensures all criteria contribute evenly. While this provides interpretability and strong performance across our datasets, it may not yield optimal combinations for more complex or heterogeneous networks. A promising future direction is to reformulate these weights as learnable parameters, enabling end-to-end training in which the selection policy is directly optimized for the downstream forecasting task. Additionally, the present model selects a single, task-independent set of sensors. This approach is appropriate for SHM applications, where sensors are typically installed once and expected to serve multiple objectives. However, in scenarios where sensors can be more flexibly added or relocated, such as environmental monitoring, task-adaptive selection strategies could tailor the sensor configuration to specific downstream tasks, improving task-specific performance.

\newpage

\appendix

\section{Details on Statistical Features}
\emph{Mean ($\mu$)}: The mean value captures the central tendency of the signal:
\begin{equation}
\mu = \frac{1}{T} \sum_{t=1}^{T} x(t),
\end{equation}
where $T$ is the total number of time samples.

\emph{Maximum ($\max$)}: The maximum value corresponds to the peak  measurement observed within the time series:
\begin{equation}
\max = \max \{ x(1), x(2), \ldots, x(T) \}.
\end{equation}
This feature is useful for identifying sensors that experience  occasional high loads or unique stress levels.

\emph{Minimum ($\min$)}: The minimum value represents the lowest observed measurement:
\begin{equation}
\min = \min \{ x(1), x(2), \ldots, x(T) \}.
\end{equation}
It highlights sensors with minimal activity.

\emph{Variance ($\sigma^2$)}: Variance measures the spread of the signal around its mean, capturing the level of fluctuation:
\begin{equation}
\sigma^2 = \frac{1}{T} \sum_{t=1}^{T} (x(t) - \mu)^2,
\end{equation}
where $\mu$ is the mean of the signal. Variance provides insights into signal stability. High-variance sensors indicate fluctuating conditions, while low-variance sensors capture more stable behaviors.

\emph{Skewness ($\gamma_1$)}: Skewness quantifies the asymmetry of the signal’s distribution:
\begin{equation}
\gamma_1 = \frac{\frac{1}{T} \sum_{t=1}^{T} (x(t) - \mu)^3}{\sigma^3},
\end{equation}
where $\sigma$ is the standard deviation. Positive or negative skewness reflects whether the signal tends toward higher or lower readings, providing insights into signal bias.

\emph{Kurtosis ($\gamma_2$)}: Kurtosis measures the “peakedness” of the signal’s distribution:
\begin{equation}
\gamma_2 = \frac{\frac{1}{T} \sum_{t=1}^{T} (x(t) - \mu)^4}{\sigma^4} - 3.
\end{equation}
High kurtosis indicates   the presence of sharp peaks or outliers, while low kurtosis suggests a  flatter  distribution.

\emph{Root Mean Square (RMS)}: RMS measures  the overall magnitude of the signal:
\begin{equation}
\text{RMS} = \sqrt{\frac{1}{T} \sum_{t=1}^{T} x(t)^2}.
\end{equation}
This feature is useful for identifying sensors with with varying levels of intensity or activity.

\section{Pseudo-Codes and Algorithms}
\begin{algorithm}
\caption{Statistical and Spectral Feature Extraction for Sensor Signals}
\label{alg:feature_extraction}
\begin{algorithmic}[1]
\STATE \textbf{Input:} Time-series data from $N$ sensors: $\{ x_i(t) \}_{i=1}^N$, where \(t = 1, \dots, T\)
\STATE \textbf{Output:} Feature vectors $\{\mathbf{f}_i\}_{i=1}^N$

\FOR{each sensor $s_i$}
    \STATE \textbf{Step 1: Compute statistical features}
    \begin{itemize}
        \item $\mu_i \gets \frac{1}{T} \sum_{t=1}^{T} x_i(t)$
        \item $\max_i \gets \max \{ x_i(1), \ldots, x_i(T) \}$
        \item $\min_i \gets \min \{ x_i(1), \ldots, x_i(T) \}$
        \item $\sigma_i^2 \gets \frac{1}{T} \sum_{t=1}^{T} (x_i(t) - \mu_i)^2$
        \item $\gamma_{1,i} \gets \frac{\frac{1}{T} \sum_{t=1}^{T} (x_i(t) - \mu_i)^3}{\sigma_i^3}$
        \item $\gamma_{2,i} \gets \frac{\frac{1}{T} \sum_{t=1}^{T} (x_i(t) - \mu_i)^4}{\sigma_i^4} - 3$
        \item $\text{RMS}_i \gets \sqrt{\frac{1}{T} \sum_{t=1}^{T} x_i(t)^2}$
    \end{itemize}
    
    \STATE \textbf{Step 2: Compute spectral features}
    \begin{itemize}
        \item $X_i(f) \gets \sum_{t=1}^{T} x_i(t) e^{-j 2 \pi f t / T}$
        \item $\mathsf{MB}_i \gets [|X_i(f_1)|, |X_i(f_2)|, \dots, |X_i(f_k)|]$
        \item $f_{\text{peak},i} \gets f_{k_{\text{peak},i}}$
        \item $\mathsf{max_f}_{i} \gets \max_k |X_i(f_k)|$
    \end{itemize}
    
    \STATE \textbf{Step 3: Store all features in the feature vector:}
    \begin{itemize}
    \item  $\mathbf{f}_i \gets [\mu_i, \max_i, \min_i, \sigma_i^2, \gamma_{1,i}, \gamma_{2,i}, \text{RMS}_i, \mathsf{MB}_i, f_{\text{peak},i}, \mathsf{max_f}_{i}]$
    \end{itemize}
\ENDFOR
\end{algorithmic}
\end{algorithm}

\begin{algorithm}
\caption{Structural Feature Extraction for Graph Nodes}
\label{alg:centrality_calculation}
\begin{algorithmic}[1]
    \STATE \textbf{Input:} Sensor graph $\mathcal{G} = (\mathcal{V}, \mathcal{E}, \mathbf{A})$
    \STATE \textbf{Output:} Centrality features $\{d_i, b_i, e_i, h_i, l_i\}_{i \in \mathcal{V}}$
    
    \FOR{each sensor $v_i \in \mathcal{V}$}
        \STATE Compute Degree Centrality: $d_i \gets \text{deg}(v_i)$
        \STATE Compute Betweenness Centrality: $b_i \gets \text{betw}(v_i)$
        \STATE Compute Eigenvector Centrality: $e_i \gets \text{eig}(v_i)$
        \STATE Compute Harmonic Centrality: $h_i \gets \text{harm}(v_i)$
        \STATE Compute Norm of Localization Operator: $l_i \gets \text{loc}(v_i)$
    \ENDFOR
\end{algorithmic}
\end{algorithm}

\begin{algorithm}
\caption{TVML for Sensor Selection}
\label{alg:sensor_selection}
\begin{algorithmic}[1]
    \STATE \textbf{Input:} Sensor time-series data $\{ x_i(t) \}_{i=1}^N$, Desired number of sensors $M$, Graph $\mathcal{G} = (\mathcal{V}, \mathcal{E}, \mathbf{A})$
    \STATE \textbf{Output:} Selected sensors $\mathcal{S}^*$
    
    \STATE \textbf{Step 1: Feature Extraction}
    \STATE Call Algorithm \ref{alg:feature_extraction} to compute $\{\mathbf{f}_i\}$

    \STATE \textbf{Step 2: Clustering}
    \STATE Apply $K$-Means clustering to $\{\mathbf{f}_i\}$ to form $M$ clusters $C = \{C_1, C_2, \ldots, C_M\}$

    \STATE \textbf{Step 3: Centrality Calculation}
    \STATE Call Algorithm \ref{alg:centrality_calculation} to compute centralities $\{d_i, b_i, e_i, h_i, l_i\}$

    \STATE \textbf{Step 4: TopoScore Calculation}
    \FOR{each sensor $v_i$}
        \STATE Calculate TopoScore: $\psi_i \gets \alpha_0 d_i + \alpha_1 b_i + \alpha_2 e_i + \alpha_3 h_i + \alpha_4 l_i$
    \ENDFOR
    
    \STATE \textbf{Step 5: Representative Selection}
    \FOR{each cluster $C_k$}
        \STATE Select representative sensor: $i^* \gets \arg \max_{i \in C_k}(\psi_i)$
        \STATE Add to selected set: $\mathcal{S}^* \gets \mathcal{S}^* \cup \{v_{i^*}\}$
    \ENDFOR
    
    \STATE \textbf{Return:} Selected sensors $\mathcal{S}^*$
\end{algorithmic}
\end{algorithm}

\section{Metrics}
Precision refers to the ratio of correctly identified positive cases to all cases predicted as positive:
\begin{equation}
    \mathsf{Precision} = \frac{\text{True Positives}}{\text{True Positives} + \text{False Positives}}.
\end{equation}

Recall measures the ratio of correctly identified positive cases to all actual positive cases:
\begin{equation}
    \mathsf{Recall} = \frac{\text{True Positives}}{\text{True Positives} + \text{False Negatives}}.
\end{equation}

F1 score provides a balance between precision and recall:
\begin{equation}
    \mathsf{F1 \ Score} = 2 \cdot \frac{\mathsf{Precision} \cdot \mathsf{Recall}}{\mathsf{Precision} + \mathsf{Recall}}.
\end{equation}

Accuracy quantifies the overall proportion of correctly classified cases:
\begin{equation}
    \mathsf{Accuracy} = \frac{\text{True Positives} + \text{True Negatives}}{\text{Total Samples}}.
\end{equation}

The RMSE measures the average magnitude of reconstruction errors:
\begin{equation}
    \mathsf{RMSE} = \frac{1}{N} \sum_{j=1}^{N} \sqrt{\frac{1}{M_j} \sum_{i=1}^{M_j} (y_{ij} - \hat{y}_{ij})^2}.
\end{equation}

The MAE quantifies the average absolute reconstruction error:
\begin{equation}
    \mathsf{MAE} = \frac{1}{N} \sum_{j=1}^{N} \frac{1}{M_j} \sum_{i=1}^{M_j} |y_{ij} - \hat{y}_{ij}|.
\end{equation}

\section{Visualizations for Results}
\begin{figure}[H]
    \centering
    \includegraphics[width=\linewidth]{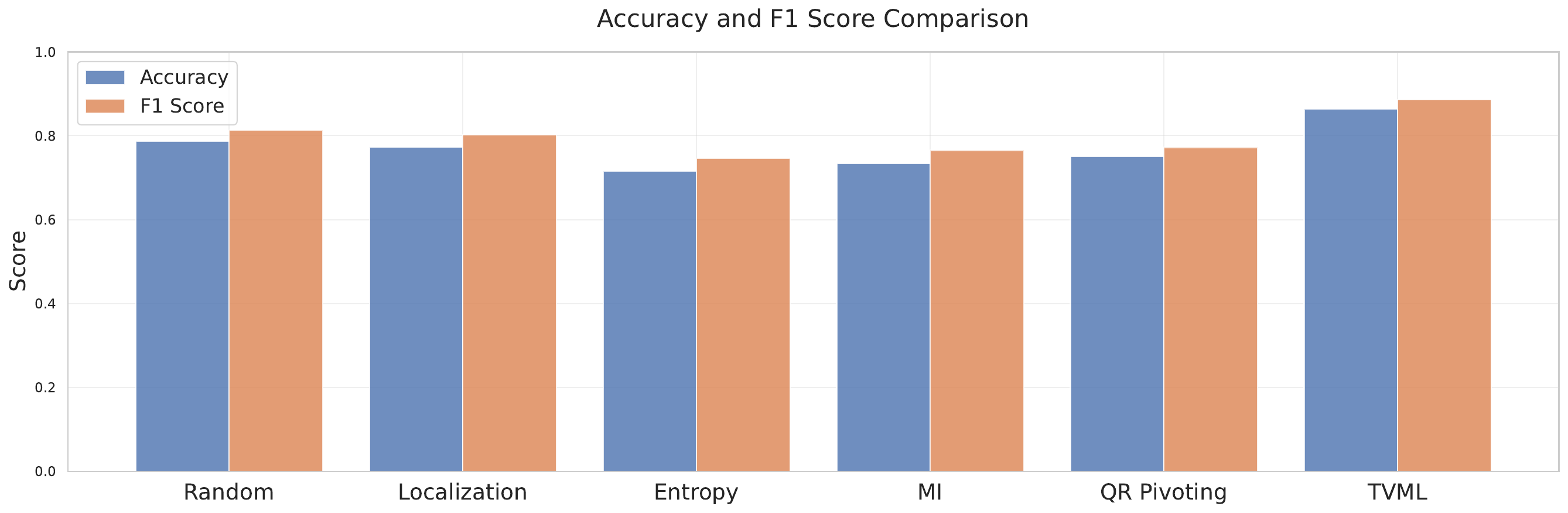}
    \caption{Comparison of Accuracy and F1 Score for Different Methods}
    \label{fig:accuracy_f1_comparison}
\end{figure}

\begin{figure}[H]
    \centering
    \includegraphics[width=0.7\linewidth]{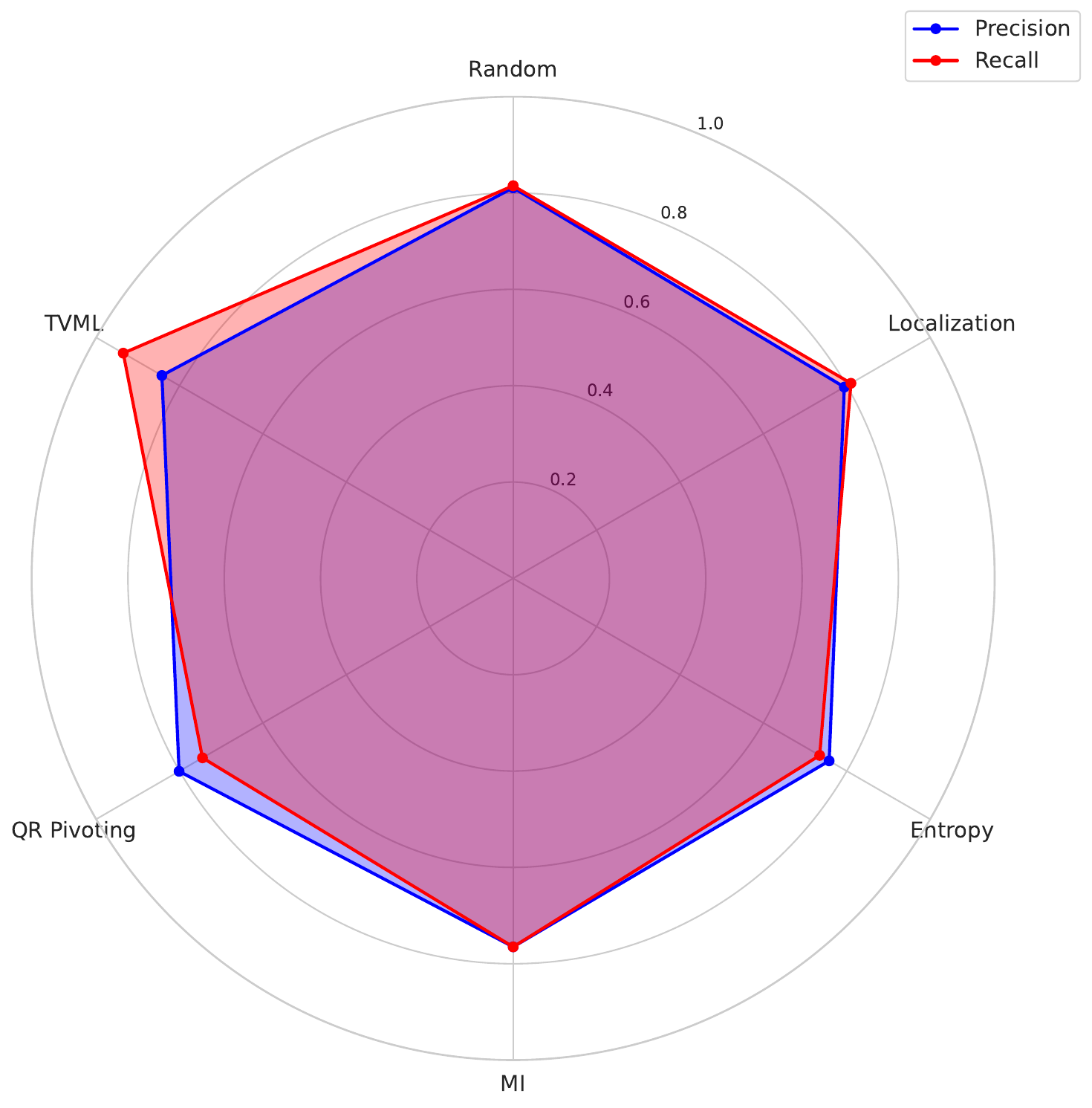}
    \caption{Precision and Recall Comparison for Different Methods}
    \label{fig:precision_recall_comparison}
\end{figure}

\textbf{Damage Detection:} Figure \ref{fig:accuracy_f1_comparison} presents a bar chart comparing the Accuracy and F1 scores across all methods, offering a clear overview of their classification performance. Accuracy measures the overall correctness of predictions, while the F1 score balances precision and recall, making it a robust indicator of a model's effectiveness. TVML significantly outperforms others in both metrics, achieving the highest accuracy of 86.4\% and F1 score of 88.6\%, demonstrating its superior ability to correctly classify instances while maintaining a balance between false positives and false negatives. Figure \ref{fig:precision_recall_comparison} provides a comparative analysis of Precision and Recall for all methods, offering insights into their relative strengths. Precision measures the accuracy of positive predictions, while Recall quantifies the ability to detect all actual positives. TVML demonstrates the highest Precision (84.2\%) and a remarkable Recall (93.5\%), making it ideal for applications where minimizing false negatives is critical.

\begin{figure}[H]
    \centering
    \includegraphics[width=1\linewidth]{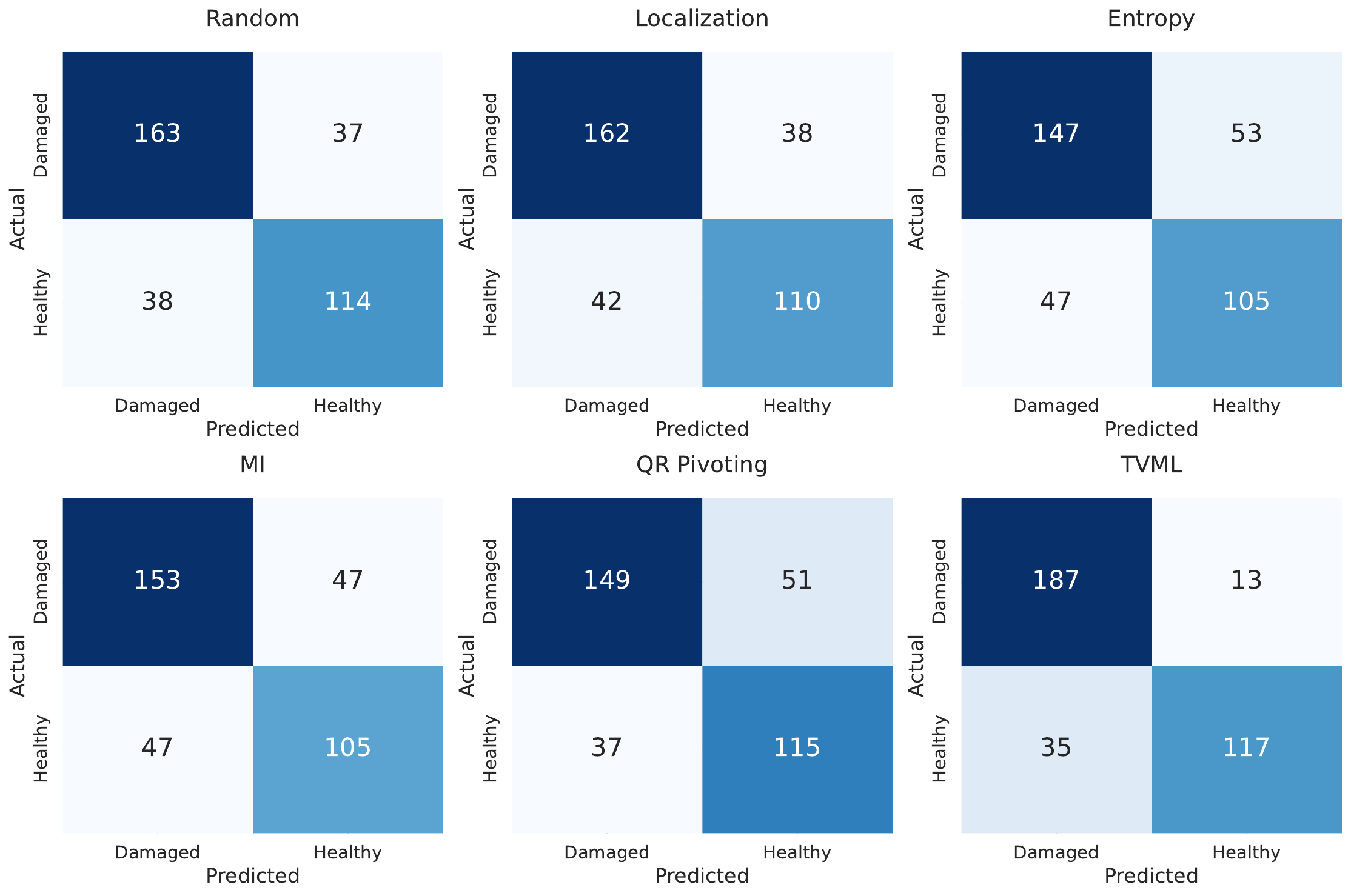}
    \caption{Confusion Matrices for Different Methods}
    \label{fig:confusion_matrices}
\end{figure}

To provide a detailed analysis of classification performance, Confusion Matrices were plotted for all methods, as shown in Figure \ref{fig:confusion_matrices}. Each matrix visually represents true positives (TP), true negatives (TN), false positives (FP), and false negatives (FN). The proposed TVML exhibits a high number of true positives (187) and very few false negatives (13), indicating its robustness in detecting positive cases accurately. In contrast, methods like Entropy, MI, and QR Pivoting show higher numbers of false negatives, reflecting a compromise in recall. The confusion matrices highlight the trade-offs each method faces between different error types and their implications for overall performance.

\begin{figure}[H]
    \centering
    \includegraphics[width=1\linewidth]{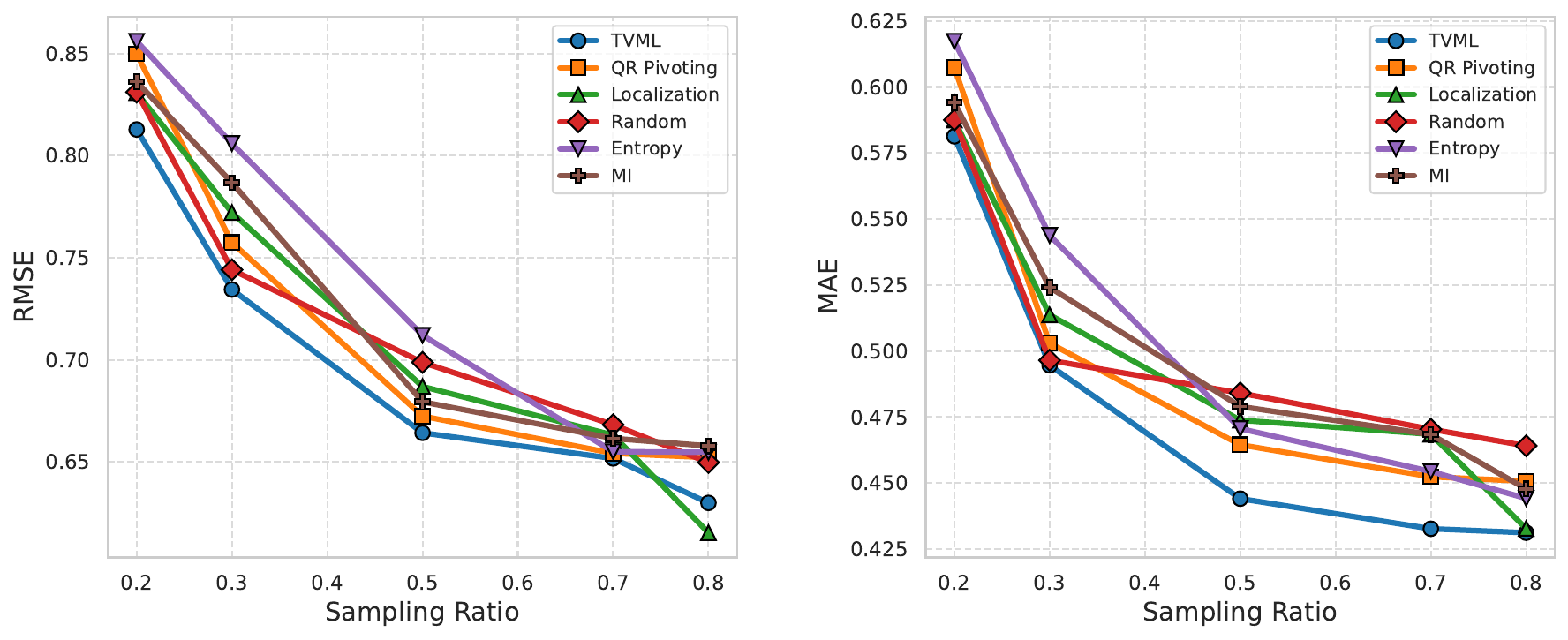}
    \caption{Comparison of Reconstructed Signals}
    \label{fig:gsr}
\end{figure}

\textbf{Time-Varying Graph Signal Reconstruction:} Figure \ref{fig:gsr} visually presents the RMSE and MAE over different sensor sets, we observe that TVML maintains a clear advantage, especially in the RMSE plot. The lines representing TVML are consistently below those of the other methods, demonstrating a lower error rate as the number of sensors increases. In both the RMSE and MAE plots, the trend suggests that as more sensors are included in the reconstruction, the performance of all methods improves, but TVML consistently outperforms other methods.

The other methods show varying performances. Entropy, MI, and Localization are competitive but do not consistently outperform TVML, with some cases where their RMSE and MAE values are closer to those of TVML. For example, Localization performs slightly better than TVML in RMSE at \( |\mathcal{S}^*| = 8 \). However, in general, TVML demonstrates its robustness and effectiveness across different cases.

\section{Feature Robustness and Dimensionality Reduction Analysis}
\label{app:feature_robustness}

Although certain statistical and spectral features may exhibit correlations, they capture complementary aspects of the sensors' temporal behavior and vibration characteristics. To evaluate whether correlated features could bias the clustering and, consequently, the sensor selection results, we conducted two complementary analyses.

First, we systematically identified highly correlated features using a pairwise correlation threshold of 0.85 and the Variance Inflation Factor (VIF), with features exceeding a VIF of 10 considered redundant. These features were removed to retain a complementary, non-redundant feature set. Clustering was then performed on both the original and reduced feature sets, and the robustness of the clustering assignments was quantitatively assessed using the Adjusted Rand Index (ARI). The high ARI value ($=1.0$) indicates that the removal of correlated features does not alter clustering assignments, confirming that feature redundancy does not introduce bias into the sensor selection process.

Second, we conducted a dimensionality reduction experiment using Principal Component Analysis (PCA). Specifically, we varied the number of retained principal components and compared the sets of sensors selected by the proposed TVML algorithm under each configuration. To quantify the similarity between the original sensor set (obtained using all features) and those obtained with reduced feature dimensions, we employed the Jaccard index, defined as:

\begin{equation}
J(A, B) = \frac{|A \cap B|}{|A \cup B|}
\end{equation}

where \( A \) and \( B \) represent two sets of selected sensors, and \( J(A, B) \in [0,1] \) measures the degree of overlap between them, with \( J(A,B)=1 \) indicating identical sets and \( J(A,B)=0 \) indicating completely disjoint selections.

\begin{figure}[h!]
    \centering
    \includegraphics[width=0.7\textwidth]{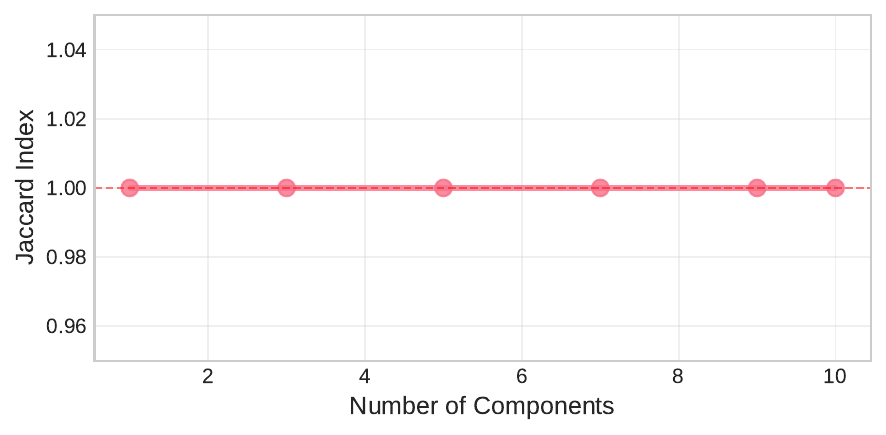}
    \caption{Jaccard similarity between the set of sensors selected using all features and the sets obtained after dimensionality reduction using different numbers of PCA components. 
    High similarity values indicate consistent sensor selection across different feature dimensions.}
    \label{fig:jaccard}
\end{figure}

As illustrated in Figure~\ref{fig:jaccard}, the Jaccard similarity remains consistently high across varying PCA dimensions, indicating that the same sensors are repeatedly selected even when the feature space is reduced. This confirms that the clustering results are stable and not significantly affected by potential feature correlations.

\section{Hyperparameters} 
To evaluate the effectiveness of sensors selected by each method, further assessment was conducted using the STGNN for both anomaly detection and time-varying graph signal reconstruction tasks. To ensure a fair and unbiased comparison, the network architecture was kept identical across all sensor selection methods. This approach prioritizes consistency in network design over hyperparameter optimization, as the primary objective is to evaluate the impact of sensor selection rather than the network's tuning. In terms of window size, a distinction is made between the two tasks. For the reconstruction task, we use an acceleration sensor, which captures rapid changes in motion and requires a smaller window size of 16 timesteps to capture these finer details. On the other hand, for the anomaly detection task, we use a displacement sensor, which monitors larger-scale movements and changes, necessitating a larger window size of 128 timesteps to capture long-term variations and trends. This difference in sensor types and the associated scale of changes justifies the variation in window sizes between the two tasks.

For the TVML method, sensor selection is governed by the TopoScore formulation in
Eq. \ref{topoeq}, which relies on a user-defined weighting vector
$\boldsymbol{\alpha} = [\alpha_0, \alpha_1, \alpha_2, \alpha_3, \alpha_4]^{\top}$
to control the relative contributions of degree, betweenness, eigenvector,
harmonic, and localization centralities. While these weights can in principle be
tuned using validation data or domain knowledge to emphasize
application-specific topological properties, in this study we adopt a uniform
weighting scheme that assumes equal contributions from all five measures. Given
the central role of the TopoScore in the TVML framework, it is therefore essential
to assess the robustness of the proposed sensor selection strategy with respect
to variations in $\boldsymbol{\alpha}$. The sensitivity of the TopoScore to the weighting vector $\boldsymbol{\alpha}$ is influenced by the connectivity structure of the underlying graph. In particular, different
graph topologies can induce varying degrees of alignment or competition among
centrality measures, which directly affects how changes in $\boldsymbol{\alpha}$
translate into sensor selection outcomes. When different centrality measures yield similar node rankings, moderate variations in the weights lead to stable performance.

To this end, we conduct a systematic sensitivity analysis by varying one
component $\alpha_i \in [0,1]$ at a time, while assigning the remaining four
weights equal values of $(1-\alpha_i)/4$. This normalization ensures that
$\boldsymbol{\alpha}$ remains a convex combination and isolates the effect of
emphasizing a single centrality measure. For each weighting configuration, the
resulting sensor set is evaluated using the downstream damage detection task,
with performance measured in terms of Accuracy and F1-score.

\begin{figure*}[t]
    \centering
    \includegraphics[width=0.95\textwidth]{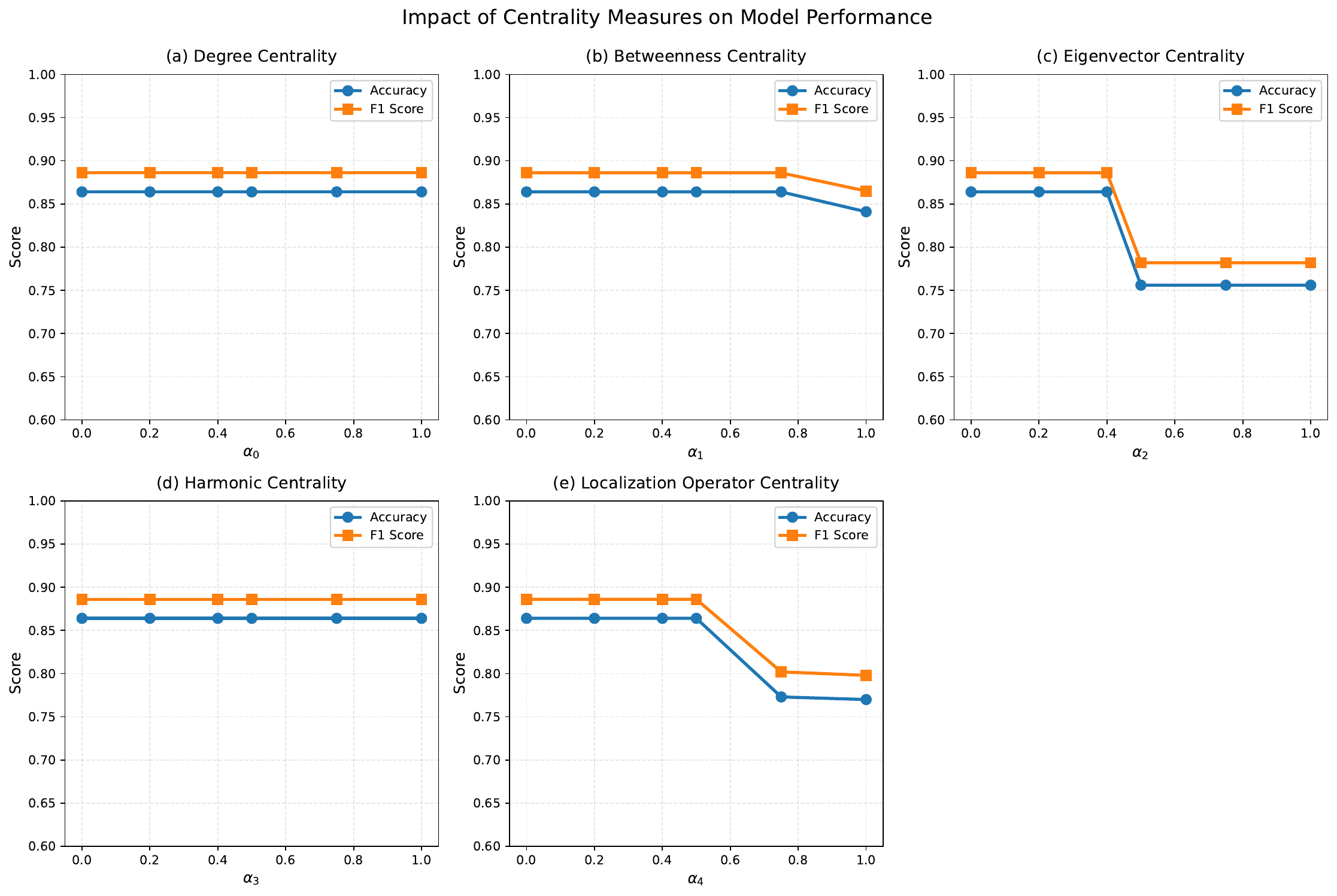}
    \caption{Sensitivity analysis of damage detection performance to weighting vector $\boldsymbol{\alpha}$. Each subplot varies a single centrality measure's weight from 0.0 to 1.0 while normalizing remaining weights uniformly: (a) Degree centrality, (b) Betweenness centrality, (c) Eigenvector centrality, (d) Harmonic centrality, and (e) Localization Operator centrality.}
    \label{fig:toposcore_sensitivity}
\end{figure*}

Figure~\ref{fig:toposcore_sensitivity} illustrates the resulting performance
trends. The results show that the TopoScore is highly robust to moderate
deviations from uniform weighting for the specific graph topology under
consideration. In particular, Degree and Harmonic
centralities exhibit complete stability across the entire weight range, yielding
identical sensor selections and unchanged performance metrics. This behavior
arises because, when the weight of either centrality is set to zero, the
remaining measures already identify sensors that possess the highest Degree and
Harmonic values within their respective clusters. As the weight of Degree or
Harmonic centrality is increased, the ranking of candidate nodes remains
unchanged, since these sensors are simultaneously favored by the other
centrality measures. Consequently, emphasizing Degree or Harmonic centrality does
not alter the selected sensor set, resulting in fully stable downstream
performance.

Betweenness centrality demonstrates near-complete robustness over most of the
weight range, with only a modest degradation when it fully dominates the
weighting vector. Specifically, when $\alpha_1 = 1.0$, Accuracy and F1-score
decrease by less than 3\%, indicating limited sensitivity to moderate
overemphasis of this measure.

In contrast, Eigenvector and Localization centralities exhibit higher sensitivity
when excessively weighted, leading to performance degradations of approximately
10--12\% when $\alpha_i = 1.0$. This behavior confirms that extreme weighting
schemes can bias sensor selection toward overly global or overly localized
structural features, respectively.

Overall, this sensitivity analysis demonstrates that moderate perturbations
around the uniform weighting yield consistent and reliable performance, while
extreme weighting configurations may adversely affect sensor representativeness.
These findings empirically justify the adoption of the uniform weighting
$\boldsymbol{\alpha} = [0.2, 0.2, 0.2, 0.2, 0.2]^\mathsf{T}$, which balances complementary
topological perspectives and avoids over-reliance on any single centrality
measure.

A comprehensive summary of the hyperparameters used in the experiments is presented in Table \ref{tab:hyperparameters}.

\begin{table}[h!]
\centering
\caption{Hyperparameters}
\resizebox{\textwidth}{!}{%
\begin{tabular}{ll}
\toprule
\textbf{Hyperparameter} & \textbf{Value} \\ \midrule\midrule
\multicolumn{2}{c}{\textbf{For STGNN}} \\ \midrule\midrule
Window Size & 16 (Reconstruction), 128 (Anomaly Detection) \\ \midrule
Activation Functions & $\mathsf{LeakyReLU}$ \\ \midrule
Number of Convolution Layers & 6 (3 for Encoder, 3 for Decoder) \\ \midrule
Output Channels for Convolution Layers & [4, 16, 16, 16, 16, 4] \\ \midrule
Number of GNN Layers & 2 (1 for Encoder, 1 for Decoder) \\ \midrule
Output Channels for GNN Layers & [1, 1] \\ \midrule
GNN Kernel Size & 2 \\ \midrule
Number of LSTM Layers & 2 (1 for Encoder, 1 for Decoder) \\ \midrule
Hidden Size for LSTM Layers & [2, 160] \\ \midrule
Batch Size & 64 \\ \midrule
Reconstruction Loss & Mean Squared Error \\ \midrule
Optimizer & Adam \\ \midrule
Learning Rate & 0.003 \\ \midrule
Patience for Early Stopping & 30 \\ \midrule\midrule

\multicolumn{2}{c}{\textbf{For TVML}} \\ \midrule\midrule
$\alpha_0, \alpha_1, \alpha_2, \alpha_3, \alpha_4$ & [0.2, 0.2, 0.2, 0.2, 0.2] \\ \midrule
Kernel for Localization Operator & $g{(\lambda_ \ell)} =e ^{-10 \frac{\lambda_ \ell}{\lambda_{\text{max}}}}$ \\ \midrule
$\lambda_1$ and $\lambda_2$ (for Graph Learning) & 0.01, 0.5 \\ \bottomrule
\end{tabular}
}
\label{tab:hyperparameters}
\end{table}

\section*{Acknowledgement}
This research was supported by the Swiss Federal Institute of Metrology (METAS).

 \bibliographystyle{elsarticle-num} 
 \bibliography{main}

\end{document}